\documentclass[preprint,3p,times,authoryear]{elsarticle}

\usepackage{amssymb} 
\usepackage{amsmath} 
\usepackage{lineno}
\usepackage{xcolor}  
\usepackage[colorlinks=true, linkcolor=black, citecolor=black, urlcolor=green]{hyperref}
\usepackage{cleveref} 
\usepackage{booktabs} 
\usepackage{xspace} 
\newcommand{\eg}{\emph{e.g.}\@\xspace}
\newcommand{\ie}{\emph{i.e.}\@\xspace}
\newcommand{\cf}{\emph{cf.}\@\xspace}

\usepackage{float} 

\journal{Journal of \LaTeX\ Templates}
\begin{document}
\begin{frontmatter}

\title{
Dense 3D Displacement Estimation for Landslide Monitoring via Fusion of TLS Point Clouds and Embedded RGB Images
} 

\author[a]{Zhaoyi Wang\corref{cor}}
\ead{zhaoyi.wang@geod.baug.ethz.ch}
\author[a,b]{Jemil Avers Butt}
\author[a]{Shengyu Huang}
\author[a]{Tomislav Medić}
\author[a]{Andreas Wieser}

\cortext[cor]{Corresponding author}
           
\address[a]{ETH Zurich, Institute of Geodesy and Photogrammetry, 8093 Zurich, Switzerland}
\address[b]{Atlas optimization GmbH, 8049 Zurich, Switzerland}

\begin{abstract}
Landslide monitoring is essential for understanding geohazards and mitigating associated risks. Existing point cloud-based methods, however, typically rely on either geometric or radiometric information and often yield sparse or non-3D displacement estimates.
In this paper, we propose a hierarchical partitioning-based coarse-to-fine approach that integrates 3D point clouds and co-registered RGB images to estimate dense 3D displacement vector fields. Patch-level matches are constructed using both 3D geometry and 2D image features, refined via geometric consistency checks, and followed by rigid transformation estimation per match.
Experimental results on two real-world landslide datasets demonstrate that the proposed method produces 3D displacement estimates with high spatial coverage (79\% and 97\%) and accuracy. 
Deviations in displacement magnitude with respect to external measurements (total station or GNSS observations) are 0.15 m and 0.25 m on the two datasets, respectively, and only 0.07 m and 0.20 m compared to manually derived references, all below the mean scan resolutions (0.08 m and 0.30 m). Compared with the state-of-the-art method F2S3, the proposed approach improves spatial coverage while maintaining comparable accuracy.
The proposed approach offers a practical and adaptable solution for TLS-based landslide monitoring and is extensible to other types of point clouds and monitoring tasks.
\end{abstract}

\end{frontmatter}



\section{Introduction}\label{sec:intro}

Landslides occur worldwide at local and regional scales, often posing significant geological hazards, particularly when they happen near human infrastructure and activities. These hazards can be triggered by natural processes (\eg, earthquake shaking, erosion, and rainfall), human activities (\eg, deforestation, construction, and excavation), or a combination of these factors~\citep{review_rs18,nat_landslide23,review_rs24}. Their consequences can be catastrophic; for example, 55,997 fatalities and widespread infrastructure damage were recorded in 4,862 landslide events worldwide between 2004 and 2016~\citep{report_numbers18}. Given the severity of these impacts, monitoring landslide behavior and changes over time (\ie, landslide monitoring) is essential to risk mitigation, facilitating effective risk management, early warning, and prediction~\citep{predict_review14,warning_review17}.

Landslide monitoring involves measuring and estimating displacements of landslides at specific points over time (\ie, across multiple epochs). Remote sensing techniques (RSTs) are indispensable for this task due to their non-contact nature, allowing the observation of geometric and/or radiometric changes without direct access to hazardous or inaccessible areas~\citep{rst_96,warning_review17}. RSTs can be operated across multiple platforms, including spaceborne systems such as satellite imagery~\citep{satellite_case14} and synthetic aperture radar (SAR)~\citep{sar_application_21}, airborne systems such as airborne LiDAR~\citep{ariborne_lidar11} and unmanned aerial vehicle (UAV) photogrammetry~\citep{uav_photogrammetry14}, and ground-based systems such as ground-based SAR~\citep{ground_sar10} and terrestrial laser scanning (TLS)~\citep{tls_landslide11}. These tools enhance landslide monitoring by offering different capabilities in range, accuracy, and spatial and temporal resolution. Among them, TLS stands out for its capability to rapidly acquire dense 3D point clouds (\eg, thousands of points per second) with high spatial resolution and measurement accuracy~\citep{lidar_landslide_12,tls_categories14,nat_landslide23}. 

To estimate displacements from multi-epoch point clouds (\eg, captured by TLS scanners), many approaches have been proposed and validated in different scenarios~\citep{change_vs_deform_15,nat_landslide23}. These methods can be broadly categorized based on the type of input information, and the nature of their outputs, namely displacement magnitudes along a predefined direction or full 3D displacement vectors. Traditional methods such as C2C~\citep{c2c2005}, C2M~\citep{c2m_1998}, and original M3C2~\citep{m3c2} rely on distances computed in Euclidean space to establish point-to-(point; approximated surface; averaged points) correspondences. While simple and widely used, these methods are sensitive to point density variation and surface roughness. 
More recently, deep learning-based approaches have emerged that establish point-wise correspondences in a high-dimensional feature space, 
with features extracted by pretrained neural networks. These methods exploit the strong generalizability of local patch-based networks, enabling deployment of models trained on annotated indoor datasets to unlabeled outdoor landslide scenes. While these approaches have shown promising results, only a few studies have explored this direction for point cloud-based landslide monitoring~\citep{f2s3,f2s3_v2,landslide22}. A more detailed description of methods of different categories is provided in~\Cref{sec:related_work_deform}.

On the one hand, most existing methods for estimating landslide displacements from 3D point clouds rely on a single type of information, either geometric or radiometric. While some studies~\citep{tls_uav_fusion_tgrs22,tls_uav_fusion24} integrate TLS and UAV-based photogrammetry for monitoring, such fusion is primarily at the data level --- aiming to generate denser point clouds --- while further analysis still relies solely on a single modality, such as geometry. 
However, using one modality may fail to estimate accurate displacements in certain scenarios that could be recovered by the other. For instance, radiometry-based approaches have been shown to perform well in regions characterized by planar or repetitive geometric structures, where the presence of radiometric texture is evident~\citep{wang2025cross}. Conversely, geometry-based methods have difficulty in achieving accurate estimation~\citep{rgb_based25}.

On the other hand, many TLS scanners contain built-in, calibrated RGB cameras that produce high-resolution RGB images, along with manufacturer-provided parameters for co-registration with the point clouds. However, these images are primarily used for visualization~\citep{visualize23} rather than for supporting displacement estimation. In our previous work~\citep{rgb_based25}, we explored the feasibility of using radiometric information from these RGB images to establish correspondences for displacement estimation, and found that using radiometric data can complement geometry-based methods. Given these findings, we argue that leveraging both modalities is essential for estimating displacement vectors with high spatial coverage and accuracy in different unfavorable situations, \eg, geometric smoothness and high illumination changes. 

To address the aforementioned issues, we propose a deep learning-based, cross-modal approach for estimating dense 3D displacements in TLS-based landslide monitoring. The key innovation lies in hierarchical partitioning-based two-stage matching strategy, which performs coarse patch matching followed by fine matching within each matched patch pair. This process jointly considers both 3D point cloud geometry and 2D RGB image information. Specifically, to mitigate unreliable correspondences computed purely in Euclidean space, we instead adopt feature-based correspondences that use deep features extracted by pretrained deep learning models with strong generalization~\citep{3d_pretrain21}, inspired by the strategy in F2S3~\citep{f2s3_v2}. To overcome the limitations of uniform or single-level partitioning, which can disrupt object boundaries or sacrifice spatial coverage, we employ a three-level hierarchical partitioning scheme that incorporates geometric and (optionally) radiometric features. By optimizing the matching pipeline and fusing 3D point clouds and 2D RGB information, our approach achieves displacement estimation accuracy comparable to the current state-of-the-art method (F2S3) while significantly improving spatial coverage.

The remainder of the paper is organized as follows:~\Cref{sec:related_work} provides a literature review of point cloud-based displacement estimation, as well as 2D and 3D deep learning-based feature description methods.~\Cref{sec:method} elaborates the principle of the proposed method for estimating 3D displacement vectors for landslide monitoring.~\Cref{sec:experiments} illustrates study areas and datasets, data preprocessing, and evaluation.~\Cref{sec:results} presents the experimental results and discussion. Finally,~\Cref{sec:conclusion} summarizes the paper and outlines future research directions. 

\section{Related work}\label{sec:related_work}

\subsection{Point cloud-based displacement estimation}
\label{sec:related_work_deform}

Many methods have been proposed to estimate displacements from point clouds acquired at different epochs, either from TLS scanners or from photogrammetric 3D reconstructions. Here we broadly classify these methods into five categories. \textbf{Category I} includes traditional methods such as C2C~\citep{c2c2005}, C2M~\citep{c2m_1998}, and original M3C2~\citep{m3c2}, which estimate displacements by comparing spatial distances in Euclidean space, typically along surface normals.
These methods operate via nearest neighbor (NN) search or projection onto local reference surfaces, using strategies such as point-to-point, point-to-mesh, or averaged point-to-neighborhood comparisons. While effective in many cases, they are sensitive to variations in point density and surface roughness and often miss tangential displacements occurring along the surface.
Several adaptations~\citet{m3c2-pm,corres_driven_m3c2,patch_m3c2} aim to improve correspondence quality or enhance the selection of informative directional estimates. Multi-directional M3C2~\citep{multi_direct_m3c2} improves flexibility by considering multiple measurement directions. 
However, without explicit point-wise correspondence tracking, these methods estimate displacement only along predefined direction(s) rather than recovering full 3D displacement vectors.

\textbf{Category II} entails ``Piecewise ICP" and related alternatives~\citep{early_dvfs07,emphraim16,icprox18}, which uniformly partition point clouds into tiles and estimate per-tile displacements as the norms of translation vectors obtained through ICP algorithm~\citep{point2point_icp92,point2plane_icp92,point2plane_icp2_96} or some of its variants. In this case, the per-tile correspondences are also established directly in the Euclidean space. Despite their effectiveness in certain scenarios, this class of methods yields biased estimates in instances where per-tile rigid body motion is violated or points within the tile are inadequate for accurate estimation.

\textbf{Category III} consists of methods that estimate 3D displacement vectors by establishing per-point correspondences using deep learning-based 3D point feature descriptors, such as F2S3~\citep{f2s3,f2s3_v2}. 
\textbf{Category IV} relies on image-based representations of the acquired point clouds, typically through hillshade rendering, or directly using RGB images~\citep{rgb_based25} to establish correspondences in 2D image space. The estimations of these methods are typically based on image correlation, optical flow, or feature matching~\citep{fey2015deriving,holst2021increasing,sattelite_img_22}.

\textbf{Category V} encompasses methods developed for point clouds generated through photogrammetric 3D reconstruction rather than TLS. For instance,~\citet{ts_rgbd} employs a total station equipped with a telescope camera to obtain 3D observations and RGB images simultaneously. Matching is then conducted on RGB images using cross-correlation methods. \citet{sift_and_reconstruction_extend} applies SIFT~\citep{sift} to UAV images to identify 2D matches, which are then reconstructed into 3D displacements using photogrammetric techniques. Similarly,~\citet{icepy4d,icepy4d_extend} perform dense 3D reconstruction from UAV data for glacier monitoring. They utilize deep learning-based image matching algorithms for matching, leveraging deep learning-based image matching to yield denser displacement vectors than earlier approaches. While these methods are relatively easy to implement, their accuracy heavily depends on reconstruction quality and they are often sensitive to large illumination variations and seasonal changes.

\subsection{2D, 3D deep learning-based feature description}

\paragraph{\textnormal{\textbf{Detector-based feature description methods}}}
Non-learning-based 2D pixel-wise matching algorithms, such as SIFT~\citep{sift}, are constrained by their sensitivity to large illumination changes and viewpoint variations~\citep{sun2021loftr}. To address these challenges, a number of studies have proposed the utilization of deep learning networks for pixel-wise matching~\citep{d2net_19,r2d2_19,aslfeat_20}.
Inspired by these works, \citet{bai2020d3feat} propose using joint feature detection and description for 3D point-wise matching. \citet{predator2021} introduce a method that predicts keypoint saliency and point cloud overlap scores to enhance detection and description. This approach specifically addresses the challenge of aligning point clouds with low overlap. Although these detector-based methods have improved the performance of pixel-wise or point-wise matching, they often struggle when keypoint detection is ineffective. Consequently, inaccurate keypoint detection can also limit the performance of feature descriptors.

\paragraph{\textnormal{\textbf{Detector-free feature description methods}}} 
Instead of explicitly detecting and describing keypoints and then matching them, detector-free methods integrate feature matching directly into the learning process. These methods typically employ a coarse-to-fine step to implement matching~\citep{li20dualrc,ZhouCVPRpatch2pix}.
More recently, LoFTR~\citep{sun2021loftr} and its variant Efficient LoFTR~\citep{wang2024eloftr}, introduce transformer-based architectures to improve the accuracy of 2D image matching.
Inspired by the coarse-to-fine matching mechanism used for 2D image matching, CoFiNet~\citep{yu2021cofinet} adopts a similar mechanism for the 3D point matching task. Building on this framework, GeoTransformer~\citep{geotrans22} incorporates transformer networks to further improve registration accuracy, particularly in challenging low-overlap scenarios. 

Our proposed method is detector-free and introduces two key innovations over existing detector-free approaches: (i) Existing methods rely on fully convolutional networks that process entire point clouds, which limits their applicability to large-scale datasets. In contrast, our method leverages a local patch-based network~\citep{3d_pretrain21} with a strong generalizability. This design is especially suited for landslide monitoring, where point clouds may contain tens of millions of points and labeled training data are scarce. (ii) Existing methods often employ uniform subsampling to generate hierarchical keypoints or patches~\citep{thomas2019KPConv}. Instead, we adopt a hierarchical, feature-based partitioning strategy that aligns better with object boundaries, enabling more effective motion pattern estimation.

\section{Methodology}\label{sec:method}

\begin{figure}[t!]
  \centering
  \includegraphics[width=0.7\linewidth]{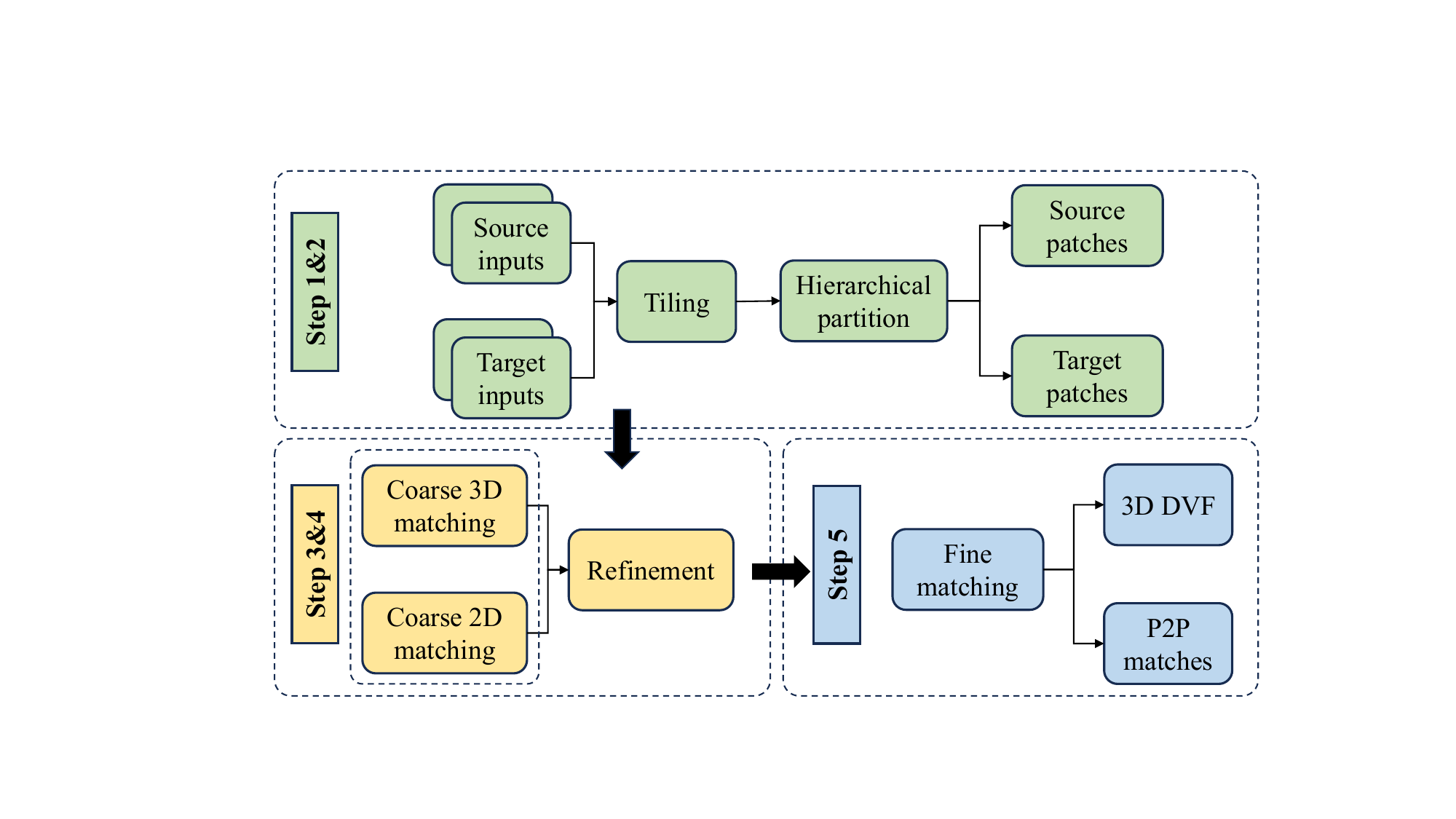}
  \vspace{-1em}
  \caption{Method overview. The proposed pipeline consists of five steps: (1) Point cloud tiles and the corresponding image tiles are generated. (2) Each paired source--target point cloud tile is partitioned into smaller patches. (3) Coarse matching establishes patch-wise correspondences / matches for each partition level based on both 3D point cloud and 2D RGB information. (4) A refinement module evaluates match quality and discards low-quality patch matches based on predefined criteria. (5) For each pair of patch matches, a rigid transformation is estimated and applied to all points within the source patch. The transformed source points and associated displacement vectors are then aggregated to construct 3D DVF, optionally producing point-to-point (P2P) matches. The final 3D DVF and P2P matches are integrated from different partition levels according to an integration step (\cf~\Cref{sec:fine_matching}).}
\label{fig:method_overview}
\end{figure}

Given two TLS-acquired point clouds of a landslide at different epochs, denoted as the source ($\mathbf{P}$) and target ($\mathbf{Q}$) point clouds, along with their associated RGB images ($\mathbf{I}^{\mathrm{p}}$ and $\mathbf{I}^{\mathrm{q}}$) obtained from built-in calibrated cameras, our objective is to estimate 
a dense 3D displacement vector field (DVF) for landslide monitoring. Formally, by jointly leveraging both 3D geometry and 2D RGB information, we aim to compute, for each point $(x, y, z)$ in $\mathbf{P}$, a displacement vector $(\Delta X, \Delta Y, \Delta Z)$ that represents its estimated motion from $\mathbf{P}$ toward its most likely counterpart in $\mathbf{Q}$:
\vspace{-0.5em}
\begin{equation}
  \mathbf{V} = \mathcal{F}(\mathbf{P}, \mathbf{Q}, \mathbf{I}^{\mathrm{p}}, \mathbf{I}^{\mathrm{q}}),
  \label{eq:object_func}
\end{equation}
where $\mathbf{V}: \mathbb{R}^3 \ni (x, y, z) \mapsto (\Delta X, \Delta Y, \Delta Z) \in \mathbb{R}^3$ denotes the 3D DVF, and $\mathcal{F}$ is the function that estimates $\mathbf{V}$.

To estimate dense 3D DVF for landslide monitoring, we design a fusion-driven approach that jointly leverages TLS point clouds and co-registered RGB images. The RGB images, captured by built-in camera(s), are co-registered with the point cloud at the respective epoch (see details in~\Cref{sec:datasets}).
The proposed method consists of five key steps: tiling large point clouds and the corresponding RGB images (\cf~\Cref{sec:tiling}); partitioning each point cloud tile into smaller patches (\cf~\Cref{sec:partition}); generating candidate coarse matches between patches (\cf~\Cref{sec:coarse_matching}); refining these coarse matches (\cf~\Cref{sec:refine}), and performing fine matching (\cf~\Cref{sec:fine_matching}). An overview of the proposed method is illustrated in~\Cref{fig:method_overview}. While designed for two epochs, the method is easily generalizable to multi-epoch scenarios by sequentially processing pairs of epochs.

\subsection{Point cloud and image tiling}\label{sec:tiling}

Due to the large size of point clouds (\eg, tens of millions of points per epoch) representing an entire landslide, we divide the point clouds into smaller spatial tiles to improve computational efficiency and reduce memory usage. 
Following~\citet{f2s3_v2}, we first project the source and target point clouds to 2D along the coordinate axis that yields the largest bounding box in the projection.
This bounding box is then recursively subdivided along the longer edge until each tile contains fewer than a predefined number of points, \eg, 1 million\footnote{The maximum number of points per tile was set to avoid GPU memory overflow during processing. Tiles exceeding this limit are recursively subdivided. Empirically, a limit of 1 million points per tile works well on a single GeForce RTX 3090 Ti (24 GB) with an AMD Ryzen 7 5800X (8 cores). This parameter can be adjusted according to available memory capacity.}.
Each tile pair, consisting of a source and a target tile that spatially overlap the same 3D region, is processed independently of the other tile pairs in the subsequent steps of our method.
To accommodate displacements that may span tile boundaries, adjacent tiles are defined with an overlap margin equal to the maximum expected displacement (\eg, 10 m, set by a manually adjustable threshold).
To ensure computational efficiency, the RGB images are divided into smaller overlapping tiles before being processed by the image matching algorithm.

\subsection{Hierarchical point cloud partitioning}\label{sec:partition}

The local rigidity assumption is commonly applied in many applications~\citep{asrigid07,earthquake_case14,rigid_scane_flow16}. We argue that this assumption also holds for landslide cases, as small objects are more likely to move as rigid bodies. For example, individual boulders are expected to move as a single unit from one location to another, rather than splitting into separate pieces. To exploit this property, we propose partitioning the tiled point clouds into small patches, where patches correspond to those small objects or small regions undergoing locally consistent motion.
Selecting an appropriate patch size is critical for accurate displacement estimation. Oversized patches may span multiple objects with non-rigid motions, violating the local rigidity assumption, whereas overly small patches tend to be dominated by noise. 

\begin{figure}[htb!]
  \centering
  \includegraphics[width=0.95\linewidth]{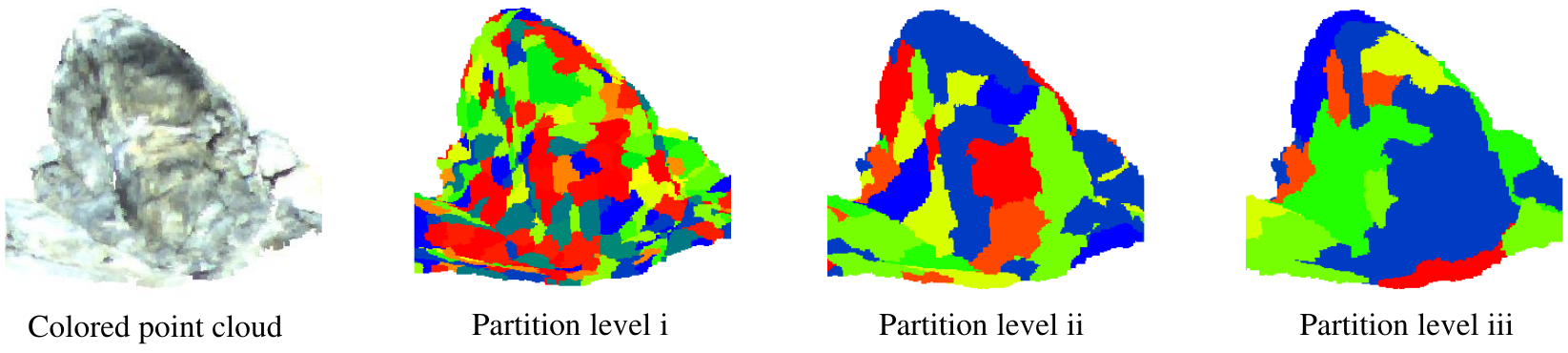}
  \vspace{-1em}
    \caption{Hierarchical partitioning of the point cloud. Starting from the RGB-colored input, the method generates three levels of partitions that capture structural boundaries at multiple scales.
    The levels are generated independently using different regularization strengths: 
    Level~i (or level 1) preserves small, geometrically homogeneous parts; Level~ii reveals larger structural components; and Level~iii highlights entire objects or coherent regions, providing global contextual information~\citep{robert2023spt}.
    }
  \label{fig:differ_partition_levels}
\end{figure}

To capture multi-scale structural variations in the scene, we adopt the hierarchical partitioning strategy, which decomposes the point cloud into three levels of granularity (\cf~\Cref{fig:differ_partition_levels}). The method relies on pre-computed geometric (linearity, planarity, and curvature) and radiometric (grayscaled RGB) features to ensure that the partitions align with both structural and appearance boundaries.
Formally, for each tiled point cloud, the partitioning process applies a sequence of graph-based optimization problems with increasing regularization parameters that control the coarseness of the resulting patches~\citep{robert2023spt}. This yields multi-scale subsets defined as:
\begin{equation}
  \mathcal{P}^{(l)} = \{ P^{(l)}_k \mid k = 1, \dots, N_l \}, \quad
  \mathcal{Q}^{(l)} = \{ Q^{(l)}_k \mid k = 1, \dots, N_l \}, \quad l \in \{1, 2, 3\},
  \label{eq:hierarchical_partition}
\end{equation}
where $P^{(l)}_k$ and $Q^{(l)}_k$ denote the $k$-th patches of the tiled source and target point clouds,
$\mathcal{P}^{(l)}$ and $\mathcal{Q}^{(l)}$, respectively, at hierarchy level $l$.

Each partition level is employed independently to guide patch-wise matching in the subsequent coarse-to-fine pipeline. At a later stage (\cf~\Cref{sec:fine_matching}), we integrate the displacement estimates from all three levels to maintain high accuracy and high spatial coverage.
To reduce noise, patches containing fewer than ten points are excluded from the matching process.


\subsection{Candidate generation for coarse matching}\label{sec:coarse_matching}

\begin{figure}[htb!]
  \centering
  \includegraphics[width=0.7\linewidth]{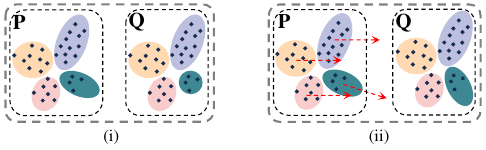}
  \vspace{-1em}
  \caption{Different patch matching cases. Each case contains four example patch matches (same color) of point cloud~$\mathbf{P}$~and~$\mathbf{Q}$. Case (i): The patch match can differ in shape (see the~\setlength{\fboxsep}{0pt}\colorbox[rgb]{0.24,0.53,0.58}{dark cyan} ones). Case (ii): The patch match can move differently compared to the neighbor patches (see the different orientations of the~\setlength{\fboxsep}{0pt}\colorbox[rgb]{1,0,0}{red} arrows).}
  \label{fig:coarse_matching_cases}
\end{figure}

For coarse matching, also known as patch matching, we define a match between a source and target patch as the pair with the highest similarity based on their geometric and radiometric attributes.
A match may still be declared even if the patches (i) differ in their shape or (ii) undergo local orientation changes relative to their neighbors, as illustrated in~\Cref{fig:coarse_matching_cases}. However, this matching procedure is impacted by the fact that the partitioning method rarely generates identical patches for the same regions across two epochs, even when the regions remain stable. To address this challenge and ensure accurate matching between patches from different epochs, we develop a processing scheme that is robust with respect to (w.r.t.) the exact patch shapes. 
This scheme contains patch matching derived from 3D source data (\cf~\Cref{sec:3d}) and patch matching derived from 2D source data (\cf~\Cref{sec:2d}). The final set of initial coarse matches combines both results, ensuring a balanced use of different types of information.

\subsubsection{Patch matching from 3D source}
\label{sec:3d}

Patch matching from the 3D source (\ie, point cloud) relies on 3D point feature matching. To ensure feature discriminativeness, we first downsample the point cloud tiles to achieve a uniform density. This prevents feature descriptors from over-focusing on densely populated regions while neglecting sparse areas. Unlike methods such as F2S3~\citep{f2s3_v2}, which use a fixed voxel size for downsampling and feature extraction across all tiles, we adjust the voxel size adaptively based on the local density (\ie, the mean scan resolution) of each tile. 
Next, we use a pretrained local patch-based 3D neural network~\citep{3d_pretrain21} to extract 3D point features from each point in the patch. Since the objective is to establish patch-wise matches, we then focus on aggregating the features of all points within each patch. 

\begin{figure}[htb!]
  \centering
  \includegraphics[width=0.95\linewidth]{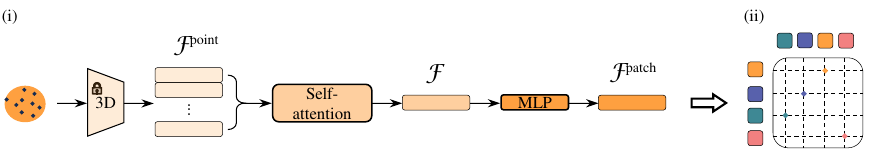}
  \vspace{-1em}
  \caption{Coarse matching based on patch feature similarity. (i) Network architecture for constructing patch features using point features extracted from a pretrained model within each patch. (ii) Mutual NN matching between patch features from the two epochs for generating candidate patch matches.}
\label{fig:coarse_3d}
\end{figure}

To achieve this, we design a neural network (\cf~\Cref{fig:coarse_3d}) that incorporates self-attention and MLP layers, and train it on publicly available indoor datasets~\citep{zeng20163dmatch}, eliminating the need for effort in annotating landslide training data. After the pretrained model assigns features to the points within a patch, our network begins by using self-attention layers~\citep{vaswani2017attention} to adaptively aggregate these features. The aggregated features then go through MLP layers to form the final patch features. Patches from the source and target point clouds are processed identically to construct patch features. During training, we apply the circle loss~\citep{sun2020circle}, a variant of the triplet loss~\citep{FCGF2019} commonly used in 3D feature matching~\citep{predator2021,geotrans22}, to optimize the patch feature aggregation. In the testing phase, \ie, when estimating displacements for the real-world landslides, patch features are extracted using our pretrained network. Patch-to-patch matches from the 3D source are then established through mutual NN matching based on their features.

\subsubsection{Patch matching from 2D source}
\label{sec:2d}

To ensure high spatial coverage of the output, we employ 2D pixel matching as a complementary approach. This approach operates on the full-resolution point clouds rather than the downsampled ones. Using the available camera intrinsic and extrinsic parameters, we first perform 2D pixel matching on the selected source and target RGB images. Then, we use a NN search to lift the established 2D pixel matches to 3D point matches, similar to the method we previously proposed in~\citet{rgb_based25}. However, point cloud tiles 
may span multiple RGB images, resulting in only a subset of points being validly projected onto any single RGB image. To address this issue, we associate each point cloud tile with its top-$k$ images, where the $k$-th image is the one with the $k$-th most projected points\footnote{For the number of images (top-$k$) used in image matching, the selection reflects a trade-off between computational efficiency and spatial coverage. In our framework, each top-$k$ selection results in $k^2$ image matching operations. We found that setting $k$ as 1 provides sufficient spatial coverage while keeping the computational cost manageable.}
Formally, for a given source point cloud tile $\mathbf{P}^t$, it is projected onto all available source images, and the top-$k$ images are selected. The same process is applied to the corresponding target point cloud tile. 
We apply image matching to the source and target images using a state-of-the-art deep learning-based image matching algorithm, Efficient LoFTR~\citep{wang2024eloftr}. To integrate all 2D pixel matches across image pairs, we first select matches from the image pair with the most matches, adding matches from additional pairs only if the corresponding 3D points have not already been matched.

\begin{figure}[htb!]
  \centering
  \includegraphics[width=0.6\linewidth]{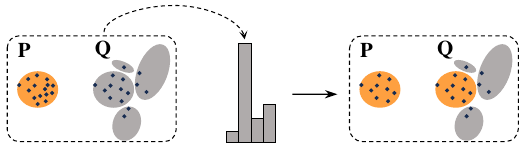}
  \vspace{-1em}
  \caption{Coarse matching based on statistic analysis.}
\label{fig:coarse_2d}
\end{figure}

Finally, we filter the lifted 3D point matches using a maximum displacement magnitude threshold, \eg, 10 m, to remove outliers with excessive displacement magnitudes.
To construct patch matches, for each source patch, we count the number of matched target points falling within each target patch. The target patch with the highest count is considered the best match for the current source patch~(see \Cref{fig:coarse_2d}).

\subsection{Patch matching refinement}\label{sec:refine}

\begin{figure}[htb!]
  \centering
  \includegraphics[width=0.4\linewidth]{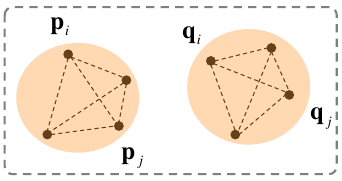}
  \vspace{-1em}
  \caption{Patch matching refinement. To check the quality of a patch match, a MADD criterion metric is computed that compares the mean pair distance between all pairs of two epochs, without the estimation of transformation.}
  \label{fig:refine}
\end{figure}

The complete set of matches $\mathbf{M}$ extracted from both the 3D source and 2D source is given as:
\vspace{-0.5em}
\begin{equation}
\mathbf{M}^{(l)} 
= \mathbf{M}^{(l)}_{3D} \cup \mathbf{M}^{(l)}_{2D}
= \big\{\, (P^{(l)}_i,\; Q^{(l)}_{m(i)}) \;\big|\; i \in \mathcal{I},\; m:\mathcal{I}\!\to\!\mathcal{J}\ \text{is injective} \,\big\},
\label{eq:initial_patch_match}
\end{equation}
where $\mathbf{M}_{3D}^{(l)}$ and $\mathbf{M}_{2D}^{(l)}$ denote the patch matches established from the 3D and 2D sources at hierarchy level $l$, respectively. $P^{(l)}_i$ and $Q^{(l)}_{m(i)}$ represent the $i$-th source patch and its corresponding target patch, respectively~(see \Cref{fig:refine}).

We observe that mismatched patch matches will result in erroneous displacement estimates, as evidenced by the presence of outliers in the final displacement vectors. To mitigate this, we propose a refinement module that evaluates the quality of patch matches and filters out those with unreliable correspondences. Inspired by the as-isometric-as-possible constraint~\citep{isometry07,isometry08} in computer graphics, our refinement module enforces local geometric consistency by preserving distances between points in the source and target patches. We introduce the Mean Absolute Distance Deviation (MADD) (\cf~\Cref{eq:madd}) as a measure of isometry preservation. A patch match is considered high quality if its MADD is below a threshold $\delta_1$, and the proportion of absolute distance deviations below $\delta_1$ exceeds a threshold $\delta_2$. 
These two thresholds are manually fine-tuned based on the characteristics of the dataset being processed (see~\ref{sec:madd_sensitive}). Only matches satisfying these criteria are retained for subsequent fine matching.
\begin{equation}\label{eq:madd}
    \mathrm{MADD} = \frac{2}{N(N-1)} 
    \sum_{0 \le i < j < N}
    \left| \| \mathbf{p}_i - \mathbf{p}_j \|_2 - \| \mathbf{q}_i - \mathbf{q}_j \|_2 \right|
\end{equation}
Here, $N$ denotes the number of correspondences. A high MADD value indicates significant deviation in the pairwise distances between corresponding points, reflecting distortion of their relative spatial structure and suggesting a poor-quality patch match. Conversely, a low MADD value implies near-isometric consistency of structure, typically associated with a rigid transformation. By enforcing this criterion, we retain only geometrically consistent matches, thereby enhancing the reliability of displacement estimation.

\subsection{Fine matching}\label{sec:fine_matching}

Previous methods~\citep{early_dvfs07,emphraim16} estimate transformations by directly applying ICP between two-epoch point clouds or their tiles. However, we observe that this process often yields inaccurate results. This is because ICP and its variants~\citep{icp_compare01} rely on Euclidean distance comparisons, making them highly sensitive to noise and large transformations. To mitigate this issue, we incorporate an initial transformation estimation step that leverages our feature-based initial matches. Additionally, to ensure accurate estimation, even when matched patches exhibit diverse shapes, we utilize only point matches rather than all points within the matched patches.

Formally, for the patch matches $\mathbf{M}_j^{(l)}$, we employ the Kabsch algorithm~\citep{kabsch} to estimate an initial rigid transformation $(\mathbf{R}_j, \mathbf{T}_j)$. This estimation is based on point matches $\{(\mathbf{p}_i,\mathbf{q}_i)\}_{i=1}^n$ within $\mathbf{M}_j^{(l)}$. The point matches are derived from the same sources used for patch matching, thereby eliminating the need for additional feature extraction. This transformation is subsequently refined using point-to-point ICP~\citep{point2point_icp92}, resulting in a refined transformation $(\hat{\mathbf{R}}_j, \hat{\mathbf{T}}_j)$. Although point matches are already relatively dense compared to most existing methods, to generate a denser output, we apply the refined transformation to all source points $\mathbf{p}_i^0$ within the current patch match. We compute the 3D DVF restricted to the current patch j as:
\begin{equation} 
\mathbf{V}_j^{(l)} : \mathbf{p}_i^0 \mapsto (\hat{\mathbf{R}}_j \cdot \mathbf{p}_i^0 + \hat{\mathbf{T}}_j - \mathbf{p}_i^0), 
\end{equation} 
where $i$ ranges through all matched points in patch $j$ and $\mathbf{V}_j^{(l)}: (x, y, z) \mapsto (\Delta X, \Delta Y, \Delta Z)$ represents the per-point displacements in the current patch match.
Finally, the 3D displacement vectors from all patch matches are collected to form the 3D DVF at hierarchy level $l$:
\vspace{-0.5em}

\begin{equation}
  \mathbf{V}^{(l)} = [\mathbf{V}_1^{(l)}, ..., \mathbf{V}_m^{(l)}],
  \label{eq:3dvfs}
\end{equation}

Optionally, if exact point-wise correspondences are required, we further extract them based on the refined transformation estimated for each matched patch pair. For each transformed source point within a patch, we search for its nearest point in the corresponding target patch. To reduce mismatches due to noise or structural variations, we discard correspondences where the point-wise distance exceeds a threshold (\eg, the mean scan resolution).

As described in~\Cref{sec:partition}, we independently estimate displacements for each of the three hierarchical partition levels, followed by an integration step that consolidates the outputs. 
Let $\mathbf{V} = [\mathbf{V}_1, \dots, \mathbf{V}_m]$ denote the integrated displacement vectors, where each $\mathbf{V}_j = \mathbf{V}_j^{(l)}$ and $l \in \{1,2,3\}$ indicates the first valid hierarchy level in descending priority (from level~i to level~iii). 
Level~i, which offers the finest partitioning with the smallest and most numerous patches, is prioritized due to its strong consistency with the local rigidity assumption. 
To handle cases of insufficient spatial coverage, we complement it with levels ii and iii. Specifically, for each point, we assign the displacement from level i if a valid estimate exists; otherwise, we fall back to level ii, and then to level iii.

\section{Experiments}\label{sec:experiments}

To evaluate the proposed method, we perform experiments on two large-scale landslide datasets.~\Cref{sec:datasets} provides details about study areas and datasets, while~\Cref{sec:preprocess} and~\Cref{sec:evaluate} outline the data preprocessing and evaluation steps, respectively.

\subsection{Study areas and datasets}\label{sec:datasets}

\begin{figure}[t!]
  \centering
  \includegraphics[width=0.9\linewidth]{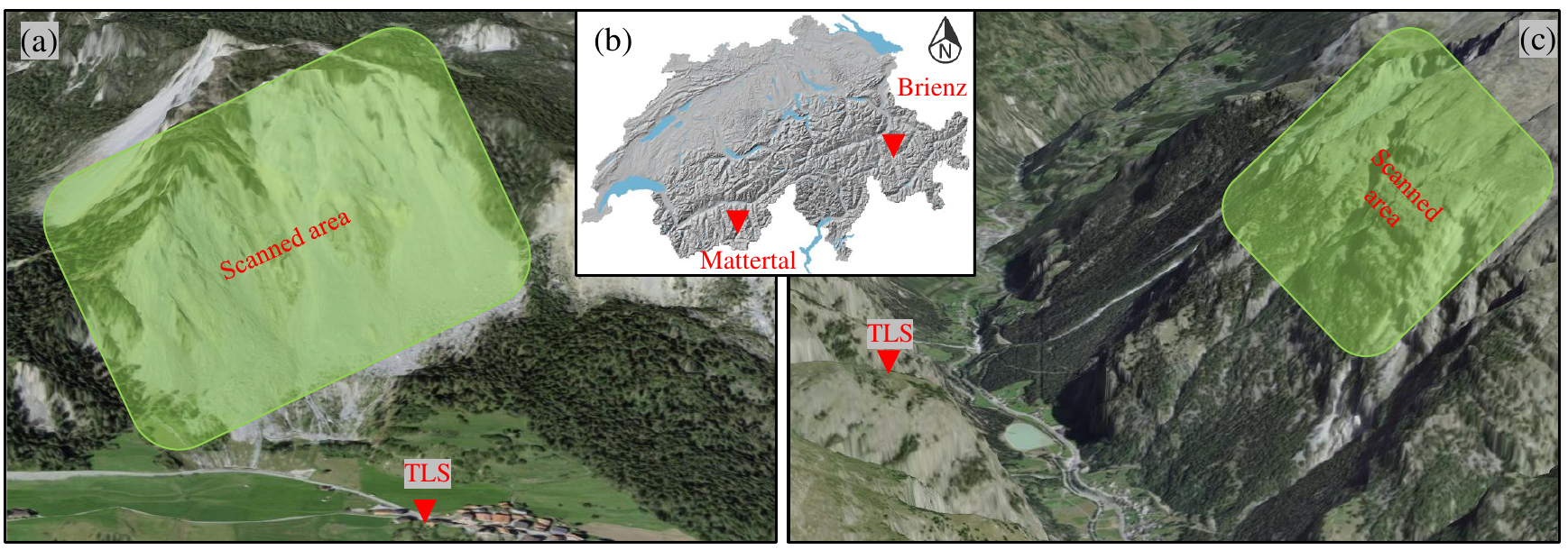}
  \vspace{-1em}
  \caption{Overview of study areas. (a): RGB image (2020) of the Brienz landslide area with the TLS scanner location indicated. (b): Map of Switzerland showing the locations of the two study areas; (c) RGB image (2019) of the Mattertal landslide area with the TLS scanner location. The two scanned landslide regions are highlighted using ~\setlength{\fboxsep}{0pt}\colorbox[rgb]{0.573,0.816,0.314}{green} rounded rectangles. Base map source: ©Swisstopo.}
  \label{fig:data_descript_both}
\end{figure}

The first study area is located north of the Brienz village in Switzerland (see~\Cref{fig:data_descript_both}(a)). This area exhibits active slope movements with displacement rates reaching several meters per year~\citep{brienz_activity_1}. These movements destabilize the upper rock mass, leading to large displacements and sporadic rockfalls. 
The Brienz TLS dataset~\citep{landslide22} was acquired using a RIEGL VZ-6000 TLS scanner positioned near the road, approximately 1,500 m from the slope. The dataset captures landslide dynamics over an area of approximately 700 $\times$ 1000 m$^2$. The scans have a mean scan resolution of 0.08 m.
RGB images were acquired using the built-in, calibrated 5-megapixel camera of the scanner. The camera captures images deflected by the laser mirror, fully covering the TLS field of view with dozens of overlapping images. Each image has a resolution of 1920 $\times$ 2560 pixels, a field of view of 5.5$^\circ$ $\times$ 7.2$^\circ$ (horizontal $\times$ vertical), a ground sampling distance (GSD) of approximately 0.05 m, and an exposure time of 5.5 ms. While manufacturer-supplied software (\eg, RiSCAN PRO v2.19.3 in our case) allows for automatic stitching into a panorama, we retain the individual images to preserve spatial accuracy and avoid stitching artifacts. For this study, we use TLS scans and their corresponding RGB images from two epochs: February and November of 2020. All images have the same intrinsic calibration but distinct poses relative to the scanner coordinate system, with parameters exported after the undistortion step in RiSCAN PRO. Additionally, observations from multiple total station (TS) prisms within the landslide area are used for quantitative evaluation. 

The second study area is located in the Matter Valley in Switzerland (see~\Cref{fig:data_descript_both}(c)). This area contains various geologically active formations, including rock glaciers and landslides, with forested terrain in the lower part and rocky terrain at higher elevations~\citep{moeller2023alpine,shi2025quantifying}. The area is subject to ongoing slope movements that could endanger local infrastructure and settlements.
The Mattertal TLS dataset was acquired using a RIEGL VZ-4000 TLS scanner positioned on stable bedrock on the opposite side of the valley, with a range of approximately 2,500 m. The dataset covers an area of approximately 3,000 $\times$ 2,000 m$^2$. The scans have a mean scan resolution of 0.30 m.
For this study, we use TLS scans and RGB images from two epochs: July 2019 and July 2022. 
Similar to the Brienz dataset, we use RGB images captured by the built-in camera. These images have the same resolution as the Brienz dataset, but they have a GSD of approximately 0.20 m and an exposure time of 11 ms. We use RiSCAN PRO to undistort the images and export the intrinsic and pose information.
Additionally, observations from multiple GNSS stations within the landslide area are used for quantitative evaluation.

\subsection{Data preprocessing}\label{sec:preprocess}

The preprocessing steps include point cloud filtering and georeferencing. Point cloud filtering removes non-ground points, \eg, those from regions covered by vegetation. These points do not directly reflect the actual displacements of the landslide surface. 
We use CloudCompare v2.13.2 for this filtering process. The next step is georeferencing, which enables comparisons with external observations, \eg, measurements from TS prisms or GNSS stations, and ensures that the point clouds from different epochs refer to the same coordinate frame.
We implement georeferencing in two different approaches according to the characteristics of each dataset. For the Brienz dataset, the landslide area is highly active.
Georeferencing was performed as part of the direct georeferencing and low-level data processing of the airborne laser scanning data by the data provider.
For the Mattertal dataset, most areas remain stable between the analyzed epochs. We align the target epoch scan to the source epoch scan using RiSCAN PRO. The registration process includes plane patch-based registration, followed by multi-station adjustment, resulting in an RMSE of 0.07 m between the two epoch point clouds. We also derive a georeferencing matrix by aligning the source epoch scan to a Swiss digital elevation model using 
a global registration pipeline (\ie, RANSAC and ICP registration) implemented in Open3D~\citep{Zhou2018}. Georeferencing is achieved by applying the georeferencing matrix to the source epoch scan and the aligned target epoch scan.

\subsection{Evaluation}\label{sec:evaluate}

The evaluation of the proposed method consists of three aspects:
(i) The accuracy of estimated displacement vectors is quantitatively assessed by comparison with external discrete observations, either from TS prisms or GNSS stations, and with manually picked points spatially closest to these observations. 
For each external observation, we identify the corresponding displacement vector by searching for its closest point in the source point cloud. This choice reflects the point-based nature of both the external observations and our outputs, enabling a direct one-to-one correspondence that avoids potential spatial smoothing. The accuracy of manual picking aligns with the mean scan resolution. Additionally, instead of searching for a single estimate, we use the mean estimated displacement vectors for comparison with external observations and manually picked points, as in existing studies~\citep{f2s3_v2,shi2025quantifying,rgb_based25}.
(ii) The spatial coverage is quantified by calculating the ratio of valid displacement estimates to the total number of points in the source point cloud for selected regions of interest (ROIs). To prevent counting duplicated estimates, both displacement estimates and source point clouds are downsampled using a voxel size corresponding to the mean scan resolution. (iii) The qualitative evaluation is performed to assess whether the spatial distribution and directionality of the estimated displacement vectors are consistent with established geomorphological understanding.

\section{Results and discussion}\label{sec:results}

In this section, we first introduce the baseline methods used for comparison in 3D displacement estimation (\Cref{sec:baseline}). Next, we evaluate the accuracy and spatial coverage in Sections~\ref{sec:quantitative_results} and~\ref{sec:qualitative_results}, respectively. The effectiveness of the proposed fusion strategy, impact of multi-level partitioning, and contribution of the refinement module are discussed in 
Sections~\ref{sec:discuss_fusion},~\ref{sec:discuss_partition}, and~\ref{sec:discuss_refine}, respectively.
A runtime analysis and discussion of method limitations and potential improvements follow in Sections~\ref{sec:runtime} and~\ref{sec:discuss_limitation}, respectively.

\subsection{Baseline selection}\label{sec:baseline}

We compare our method against three existing approaches for estimating dense 3D DVFs in landslide monitoring:
Piecewise ICP~\citep{emphraim16}, F2S3~\citep{f2s3_v2}, and our previously proposed RGB-guided method~\citep{rgb_based25}. 
These methods represent Categories II, III, and IV from~\Cref{sec:related_work_deform}, covering key approaches for dense 3D DVF estimation. 
Categories I and V are omitted because Category I does not provide full 3D vectors, and Category V is not applicable to TLS point clouds.
To enable full-scene landslide analysis, we adapt the Piecewise ICP and RGB-guided methods, which are limited to single regions of interest, by incorporating the tiling process. For other settings, we follow the choices described in the respective publications.

\subsection{Accuracy evaluation}\label{sec:quantitative_results}

We compare the estimated 3D displacements of our method with external observations: TS prisms on the Brienz dataset and GNSS stations on the Mattertal dataset. We first present the quantitative results of our method for both landslides in \Cref{tab:quantitative_external_compare}, where we use NN search to find the estimates of corresponding external observations.
Although the estimated displacement magnitudes may appear large, they fall within the range documented for the Swiss Alps in previous studies (see \ref{sec:large_displace}).
For clarity of interpretation, the coordinate system for both datasets is defined following the right-hand rule, where the Y-axis approximately corresponds to the horizontal projection of the mean line-of-sight (LoS) direction, the X-axis is perpendicular to the Y-axis horizontally, and the Z-axis points vertically upward.
The displacement magnitude deviations (\ie, $|\Delta D_S|$) between our estimates and the external observations are 0.15 m and 0.25 m, respectively. 
The larger deviation observed on the Mattertal dataset is mainly due to the fact that the mean footprint size is 0.68 m, which is larger than the 0.23 m size on the Brienz dataset. Hence, the 0.25 m deviation falls within the expected accuracy range of the Mattertal dataset.
Additionally, we manually select corresponding points at the second epoch (\ie, target epoch) to derive displacements that accurately reflect the input data used in both our method and other baseline approaches.
The deviations between our method and the manually selected reference displacements are 0.07 m and 0.20 m for the Brienz and Mattertal datasets, respectively. 
These deviations are smaller than the mean scan resolutions of 0.08 m and 0.30 m for these two datasets, respectively. This demonstrates the high accuracy of the displacement estimation by our method. Furthermore, our method has slightly larger deviations with external observations than with manual reference data, which can be attributed to georeferencing errors. To avoid this, we use the manually picked points as the reference for comparing our method with baseline methods.

\begin{table}[h]
  \centering
  \caption{Comparison of 3D displacement estimates from our method (based on NN search) against external observations and manually labeled data. 
  The upper part of the table shows results for the Brienz landslide, evaluated using TS observations and manually labeled data; the lower part presents results for the Mattertal landslide, evaluated using GNSS observations and manually labeled data. 
  $|\Delta D_X|$, $|\Delta D_Y|$, $|\Delta D_Z|$ denote the absolute deviations in the individual displacement components, while $|\Delta D_S|$ represents the absolute deviation in the displacement magnitude.}
  \label{tab:quantitative_external_compare}
  \resizebox{0.99\columnwidth}{!}{
  \begin{tabular}{ccc|cccc|cccc|cccc}
      \toprule
      & & & \multicolumn{4}{c|}{Ours (m)} & \multicolumn{4}{c|}{\(|\mathrm{Ours - External}|\) (m) $\downarrow$} & \multicolumn{4}{c}{\(|\mathrm{Ours - Manual}|\)  (m) $\downarrow$} \\
      Point & Range (km) & Footprint (m) & $D_X$ & $D_Y$ & $D_Z$ & $D_S$ & $|\Delta D_X|$ & $|\Delta D_Y|$ & $|\Delta D_Z|$ & $|\Delta D_S|$
            & $|\Delta D_X|$ & $|\Delta D_Y|$ & $|\Delta D_Z|$ & $|\Delta D_S|$ \\
      \midrule
      \multicolumn{15}{c}{Quantitative results on the Brienz dataset} \\
      \midrule
      T1 & 1.97 & 0.24 & ~0.36 & -1.84 & -4.36 & 4.75 & 0.16 & 0.04 & 0.06 & 0.19 & 0.09 & 0.08 & 0.11 & 0.13 \\
      T2 & 2.00 & 0.24 & ~0.11 & -0.21 & -0.30 & 0.38 & 0.07 & 0.03 & 0.17 & 0.23 & 0.08 & 0.15 & 0.10 & 0.17 \\
      T3 & 1.65 & 0.20 & ~0.24 & -0.38 & -0.64 & 0.78 & 0.46 & 0.05 & \textbf{0.01} & \textbf{0.00} & 0.06 & 0.15 & 0.28 & 0.12 \\
      T4 & 1.80 & 0.22 & ~0.31 & -0.68 & -0.60 & 0.96 & 0.13 & 0.01 & 0.36 & 0.28 & \textbf{0.01} & 0.08 & 0.06 & 0.02 \\
      T5 & 1.91 & 0.23 & ~0.91 & -0.94 & -1.23 & 1.80 & 0.07 & 0.05 & 0.26 & 0.05 & 0.17 & \textbf{0.03} & \textbf{0.03} & 0.05 \\
      T6 & 1.81 & 0.22 & ~0.55 & -0.82 & -0.70 & 1.21 & 0.07 & \textbf{0.00} & 0.29 & 0.16 & 0.09 & 0.14 & 0.20 & 0.03 \\
      T7 & 1.88 & 0.23 & ~1.13 & -1.54 & -1.94 & 2.72 & \textbf{0.01} & 0.07 & 0.09 & 0.16 & 0.01 & 0.04 & 0.04 & \textbf{0.00} \\
      Mean & 1.90 & 0.23 & -- & -- & -- & -- & 0.14 & 0.04 & 0.17 & 0.15 & 0.07 & 0.09 & 0.12 & 0.07 \\
      \midrule
      \multicolumn{15}{c}{Quantitative results on the Mattertal dataset} \\
      \midrule
      G1 & 4.44 & 0.67 & -0.06 & -0.08 & -0.17 & 0.20 & 0.06 & 0.08 & 0.18 & 0.19 & \textbf{0.01} & \textbf{0.02} & 0.12 & \textbf{0.07} \\
      G2 & 4.58 & 0.69 & -3.00 & -4.39 & -1.90 & 5.65 & 0.94 & \textbf{0.01} & 0.27 & 0.52 & 0.48 & 0.06 & \textbf{0.02} & 0.28 \\
      G3 & 4.62 & 0.69 & ~0.95 & -1.01 & -0.50 & 1.47 & 0.50 & 0.06 & 0.12 & 0.35 & 0.48 & 0.06 & 0.19 & 0.26 \\
      G4 & 4.40 & 0.66 & ~0.29 & -0.83 & -0.30 & 0.93 & \textbf{0.03} & 0.14 & \textbf{0.01} & \textbf{0.13} & 0.16 & 0.07 & 0.41 & 0.21 \\
      Mean & 4.51 & 0.68 & -- & -- & -- & -- & 0.37 & 0.05 & 0.10 & 0.25 & 0.28 & 0.05 & 0.19 & 0.20 \\
      \bottomrule
  \end{tabular}
  }
\end{table}

We further present the quantitative results of our method for both landslides in \Cref{tab:quantitative_external_compare_average}, where we select a set of estimated displacement vectors within a certain radius and compare their mean with corresponding external observations or manually picked points.
This comparison aligns with the evaluation used in previous studies~\citep{f2s3_v2,shi2025quantifying,rgb_based25}.
We use a 15 m radius for both datasets by default. However, for point T5, we observe a significant deviation in the mean absolute deviation (MAD) due to local non-rigid motion. Therefore, we reduce the radius to 2.5 m for point T5. 
The mean deviations between the estimated displacement magnitudes from our method and the external observations are 0.12 m and 0.30 m for the two datasets, respectively, both smaller than the corresponding footprint sizes of 0.23 m and 0.68 m.
The mean deviations between the estimated displacement magnitudes from our method and the manually selected reference displacements are 0.10 m and 0.12 m, both smaller than the mean scan resolutions of 0.08 m and 0.30 m.
These findings are consistent with our analysis based on NN search in~\Cref{sec:quantitative_results}. The MAD of the estimated displacement magnitudes are 0.15 m and 0.24 m, respectively, suggesting the high precision of the displacement magnitudes estimated by our method.

\begin{table}[h]
  \centering
  \caption{Comparison of 3D displacement estimates from our method (based on averaging) against external observations and manually labeled data. 
  The upper part of the table shows results for the Brienz landslide, evaluated using TS observations and manually labeled data; the lower part presents results for the Mattertal landslide, evaluated using GNSS observations and manually labeled data. $\overline{D_X}$, $\overline{D_Y}$, $\overline{D_Z}$,  denote the mean estimated displacement components, and $\overline{D_S}$ denotes the mean estimated displacement magnitude}
  \label{tab:quantitative_external_compare_average}
  \resizebox{0.99\columnwidth}{!}{
  \begin{tabular}{ccc|cccc|cccc|cccc|cccc}
      \toprule
      & & & \multicolumn{4}{c|}{Ours (m)} & \multicolumn{4}{c|}{\(|\mathrm{Ours - External}|\) (m) $\downarrow$} & \multicolumn{4}{c|}{\(|\mathrm{Ours - Manual}|\) (m) $\downarrow$} & \multicolumn{4}{c}{MAD (m) $\downarrow$} \\
      Point & Range (km) & Footprint (m) & $\overline{D_X}$ & $\overline{D_Y}$ & $\overline{D_Z}$ & $\overline{D_S}$ & $|\Delta D_X|$ & $|\Delta D_Y|$ & $|\Delta D_Z|$ & $|\Delta D_S|$
            & $|\Delta D_X|$ & $|\Delta D_Y|$ & $|\Delta D_Z|$ & $|\Delta D_S|$ & $\mathrm{MAD}(D_X)$ & $\mathrm{MAD}(D_Y)$ & $\mathrm{MAD}(D_Z)$ & $\mathrm{MAD}(D_S)$ \\
      \midrule
      \multicolumn{19}{c}{Quantitative results on the Brienz dataset} \\
      \midrule
      T1 & 1.97 & 0.24 & -0.08 & -1.92 & -4.09 & 4.65 & 0.28 & 0.12 & 0.32 & 0.10 & 0.35 & \textbf{0.00} & 0.38 & 0.22 & 0.53 & 0.27 & 0.38 & 0.30 \\
      T2 & 2.00 & 0.24 & ~0.18 & -0.17 & -0.30 & 0.41 & 0.15 & \textbf{0.00} & 0.17 & 0.20 & 0.16 & 0.11 & 0.09 & 0.20 & \textbf{0.04} & 0.04 & 0.05 & \textbf{0.04} \\
      T3 & 1.65 & 0.20 & -0.23 & -0.39 & -0.56 & 0.78 & 0.93 & 0.06 & \textbf{0.08} & \textbf{0.01} & 0.42 & 0.14 & 0.20 & 0.12 & 0.58 & 0.13 & 0.15 & 0.22 \\
      T4 & 1.80 & 0.22 & ~0.06 & -0.69 & -0.67 & 1.00 & 0.38 & 0.02 & 0.28 & 0.24 & 0.23 & 0.09 & 0.02 & 0.06 & 0.25 & 0.12 & 0.10 & 0.11 \\
      T5 & 1.91 & 0.23 & ~0.94 & -0.92 & -1.25 & 1.80 & 0.05 & 0.03 & 0.23 & 0.05 & 0.14 & 0.01 & 0.05 & 0.05 & 0.06 & \textbf{0.03} & \textbf{0.03} & \textbf{0.04} \\
      T6 & 1.81 & 0.22 & ~0.08 & -0.86 & -0.62 & 1.25 & 0.54 & 0.04 & 0.37 & 0.12 & 0.38 & 0.10 & 0.12 & 0.07 & 0.45 & 0.31 & 0.17 & 0.29 \\
      T7 & 1.88 & 0.23 & ~1.15 & -1.47 & -1.92 & 2.71 & \textbf{0.02} & \textbf{0.00} & 0.11 & 0.15 & \textbf{0.02} & 0.10 & \textbf{0.01} & \textbf{0.01} & 0.06 & 0.10 & 0.09 & 0.08 \\
      Mean & 1.86 & 0.23 & -- & -- & -- & -- & 0.34 & 0.04 & 0.22 & 0.12 & 0.24 & 0.08 & 0.12 & 0.10 & 0.28 & 0.14 & 0.14 & 0.15 \\
      \midrule
      \multicolumn{19}{c}{Quantitative results on the Mattertal dataset} \\
      \midrule
      G1 & 4.44 & 0.67 & -0.07 & -0.09 & -0.15 & 0.17 & \textbf{0.07} & 0.09 & 0.16 & 0.16 & 0.02 & \textbf{0.01} & \textbf{0.10} & 0.04 & \textbf{0.03} & \textbf{0.02} & \textbf{0.03} & \textbf{0.04} \\
      G2 & 4.58 & 0.69 & -2.60 & -4.22 & -1.77 & 5.73 & 0.54 & 0.19 & 0.14 & 0.61 & 0.08 & 0.12 & 0.15 & 0.37 & 0.51 & 0.18 & 0.14 & 0.23 \\
      G3 & 4.62 & 0.69 & ~0.94 & -0.98 & -0.48 & 1.24 & 0.49 & \textbf{0.02} & 0.11 & \textbf{0.12} & 0.47 & 0.10 & 0.18 & \textbf{0.03} & 0.09 & 0.13 & 0.09 & 0.27 \\
      G4 & 4.40 & 0.66 & ~0.44 & -0.72 & -0.35 & 1.11 & 0.18 & 0.03 & \textbf{0.03} & 0.30 & \textbf{0.01} & 0.04 & 0.37 & 0.04 & 0.50 & 0.30 & 0.30 & 0.40 \\
      Mean & 4.51 & 0.68 & -- & -- & -- & -- & 0.32 & 0.08 & 0.11 & 0.30 & 0.15 & 0.07 & 0.20 & 0.12 & 0.28 & 0.16 & 0.14 & 0.24 \\
      \bottomrule
  \end{tabular}
  }
\end{table}

\begin{figure}[htb!]
  \centering
  \includegraphics[width=0.7\linewidth]{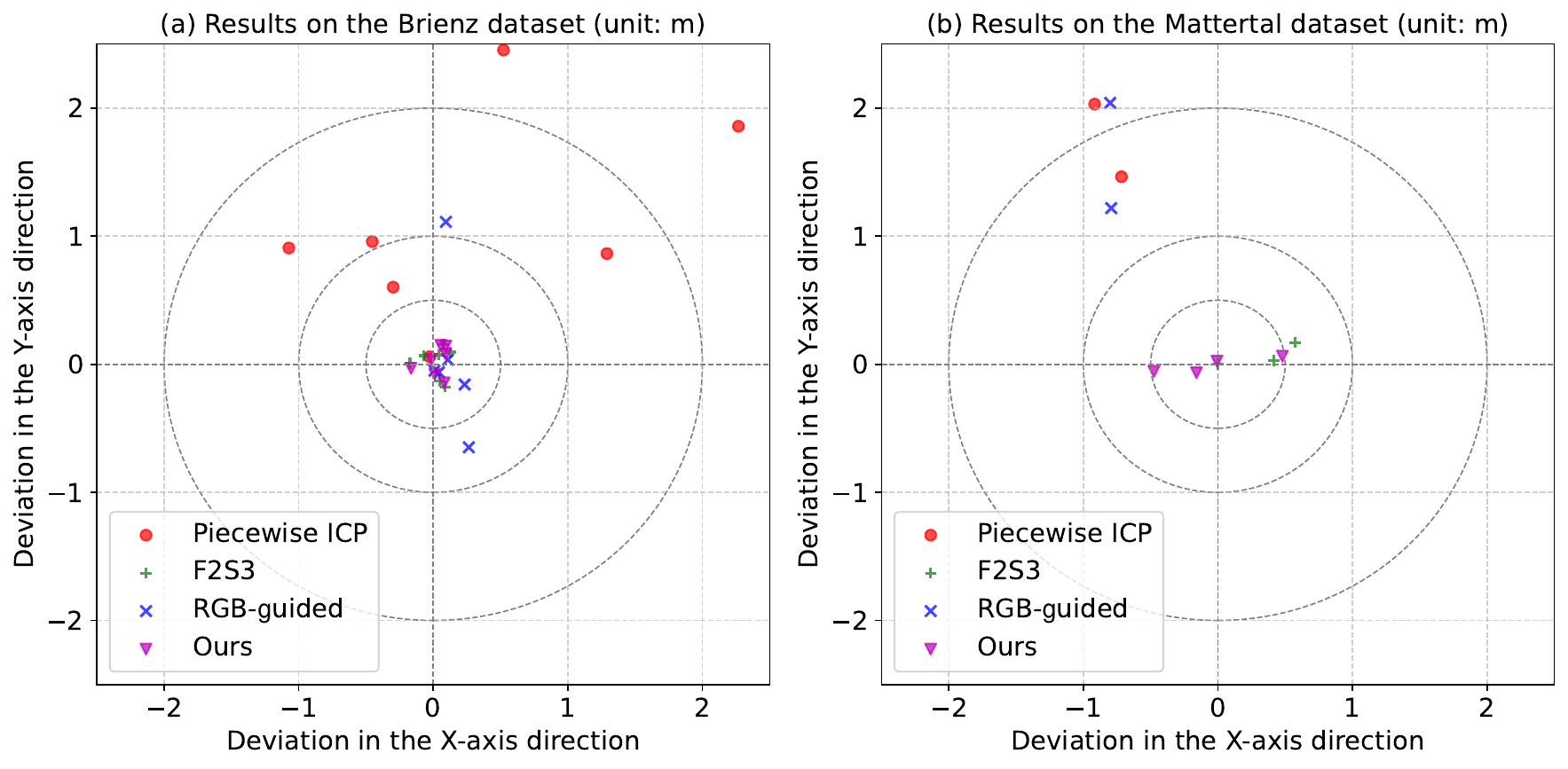}
  \vspace{-1em}
  \caption{Quantitative comparison of estimated displacements from different methods w.r.t. the manual reference, focusing on deviations along the X- and Y-axis directions (\ie, $\Delta D_X$ and $\Delta D_Y$). Dashed circles and grid lines are added to improve readability. Similar deviation patterns are observed along the X- and Z-axis directions as well (see~\ref{sec:appendix_xy_deviation}).
  }
\label{fig:quantitative_compare}
\end{figure}

Additionally, we compare the estimated displacements of our method with those of baseline methods in \Cref{fig:quantitative_compare}, where we use manually picked points as the reference and compare the deviations between the different methods and the reference.
Our method shows maximum deviations of 0.17 m along the X-axis for the Brienz dataset and 0.48 m along the X-axis for the Mattertal dataset. Among all baseline methods, F2S3 achieves the closest performance to ours, with maximum deviations of 0.18 m along the Y-axis for the Brienz dataset and 0.57 m along the Y-axis for the Mattertal dataset. All baseline methods fail to provide valid estimates corresponding to all external observations (\eg, Piecewise ICP only provides two valid estimates for all four GNSS stations on the Mattertal dataset) due to their reliance on only one type of information (either 3D point cloud or 2D RGB information). In contrast, our method provides estimates around all external observations.
Piecewise ICP~\citep{emphraim16} exhibits the largest deviations on both datasets, primarily due to its approach of uniformly partitioning and subsequently matching the closest patches based on Euclidean distance. 
The RGB-guided approach shows large deviations along the Y-axis direction. 
An investigation in \citet{assess_align} suggests that, even though the manufacturer provides co-registration between the TLS scanner and its built-in camera(s), misalignment errors of several pixels can still occur between the point clouds and the images.
These errors persist across epochs due to variations in the camera poses associated with the captured images, even when the same scanner is used. 
This is particularly problematic on the Mattertal dataset, where the mean GSD is 0.2 m, resulting in potential meter-level displacement errors.
In our approach, 2D information mainly contributes to patch matching. Due to the refinement, our method with fused 3D and 2D information still is comparable in accuracy to the 3D-based approach, F2S3.

\subsection{Spatial coverage evaluation}\label{sec:qualitative_results}

\begin{figure}[t!]
  \centering
  \includegraphics[width=0.9\linewidth]{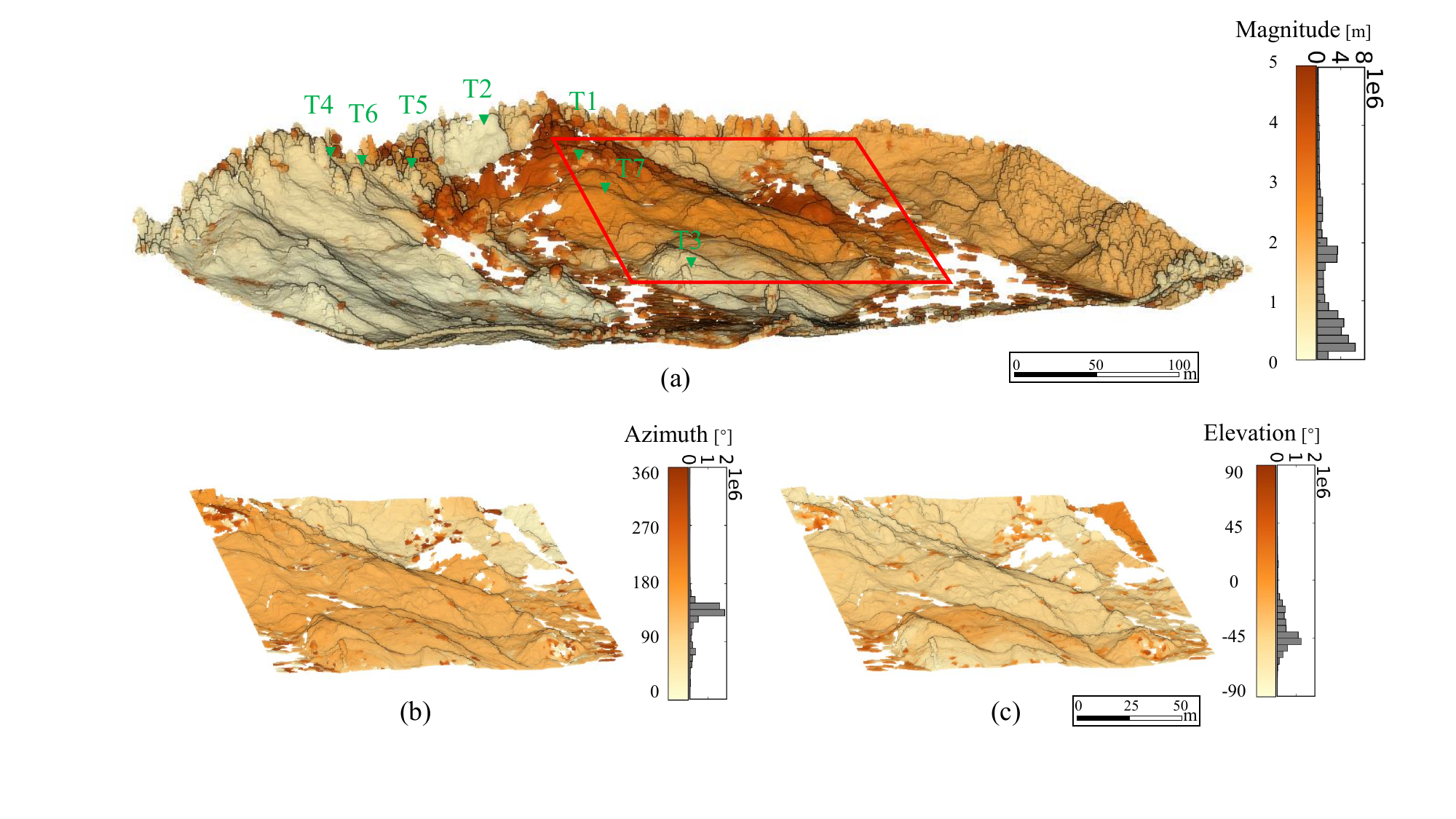}
  \vspace{-1em}
  \caption{Estimated 3D displacements on the Brienz landslide. (a): Displacement magnitudes across the entire landslide (ROI highlighted in ~\setlength{\fboxsep}{0pt}\colorbox[rgb]{1,0,0}{red}). (b): The azimuth direction of displacements within the ROI. (c): The elevation direction of displacements within the ROI.}
\label{fig:qualitative_3d_component_brienz}

\end{figure}

\begin{figure}[t!]
  \centering
  \includegraphics[width=0.8\linewidth]{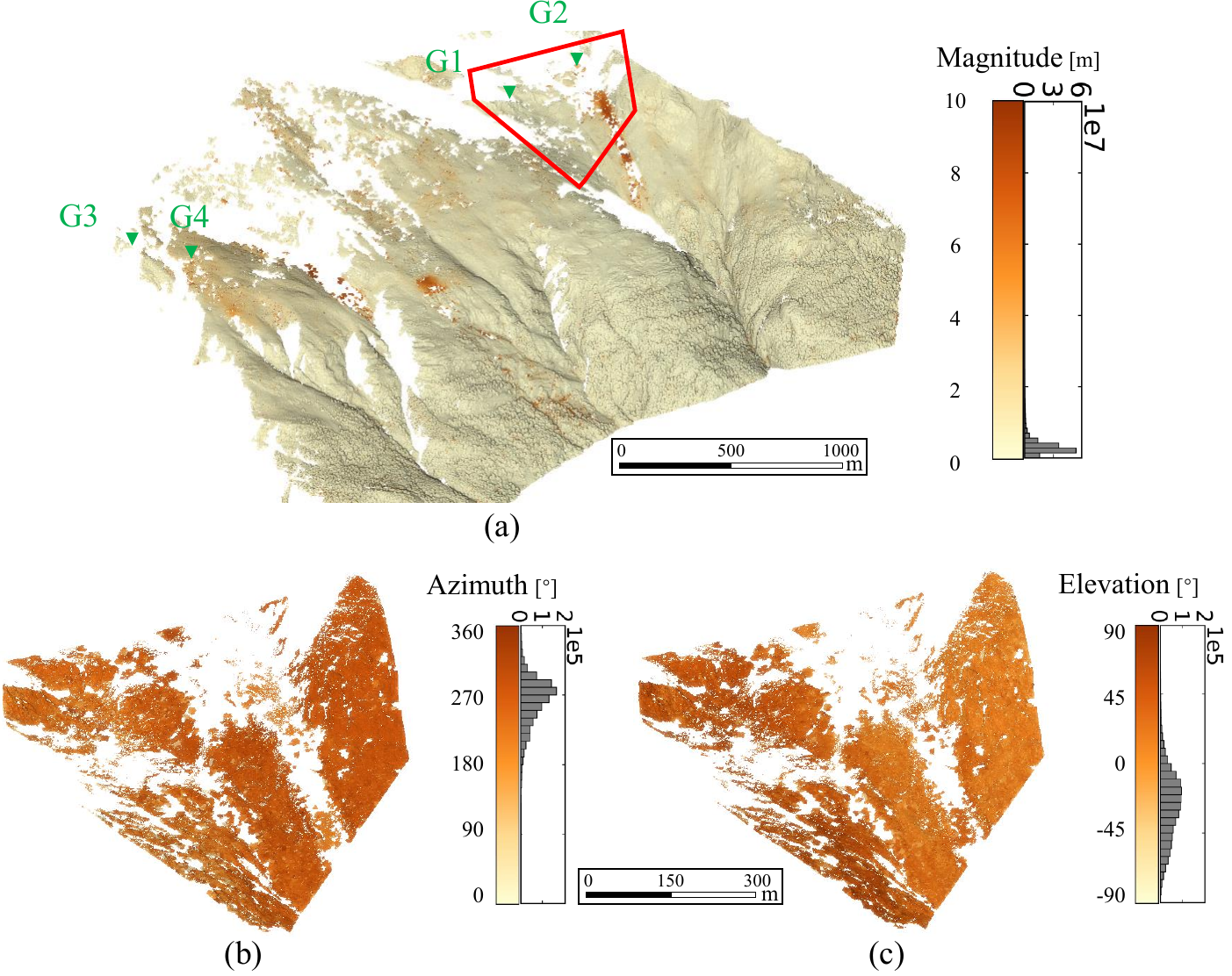}
  \vspace{-1em}
  \caption{Estimated 3D displacements on the Mattertal landslide. (a): Displacement magnitudes across the entire landslide (ROI highlighted in~\setlength{\fboxsep}{0pt}\colorbox[rgb]{1,0,0}{red}). (b) The azimuth direction of displacements within the ROI. (c): The elevation direction of displacements within the ROI.}
\label{fig:qualitative_3d_component_mattertal}

\end{figure}

We present the 3D displacements estimated by our method in \Cref{fig:qualitative_3d_component_brienz} and \Cref{fig:qualitative_3d_component_mattertal} for the Brienz and Mattertal datasets, respectively. The results demonstrate that our method can estimate 3D displacements across most landslide areas, with spatial coverage of 79\% and 97\% w.r.t. the source point cloud for the Brienz and Mattertal datasets, respectively.
The Brienz landslide exhibits significant variation in displacement magnitude across the area, whereas the Mattertal landslide remains relatively stable, with only a few localized areas showing notable movement between the observed epochs. 
To further analyze the displacement characteristics, we select a ROI from the highly active areas of each landslide, where several external observations are also included in the corresponding ROI. In the Brienz ROI (see \Cref{fig:qualitative_3d_component_brienz}), the 3D displacement vectors estimated by our method have a mean magnitude of approximately 2.5 m and mean azimuth and elevation angles of approximately 135$^\circ$ and -45$^\circ$, respectively.
This suggests displacement toward the southeast.
In contrast, most of the displacement magnitudes in the Mattertal landslide ROI (see \Cref{fig:qualitative_3d_component_mattertal}) are close to zero, and the azimuth and elevation angles exhibit large spreads without consistent direction trends. This indicates a lack of coherent surface motion during the observed period.

\begin{figure}[htb!]
  \centering
  \includegraphics[width=0.9\linewidth]{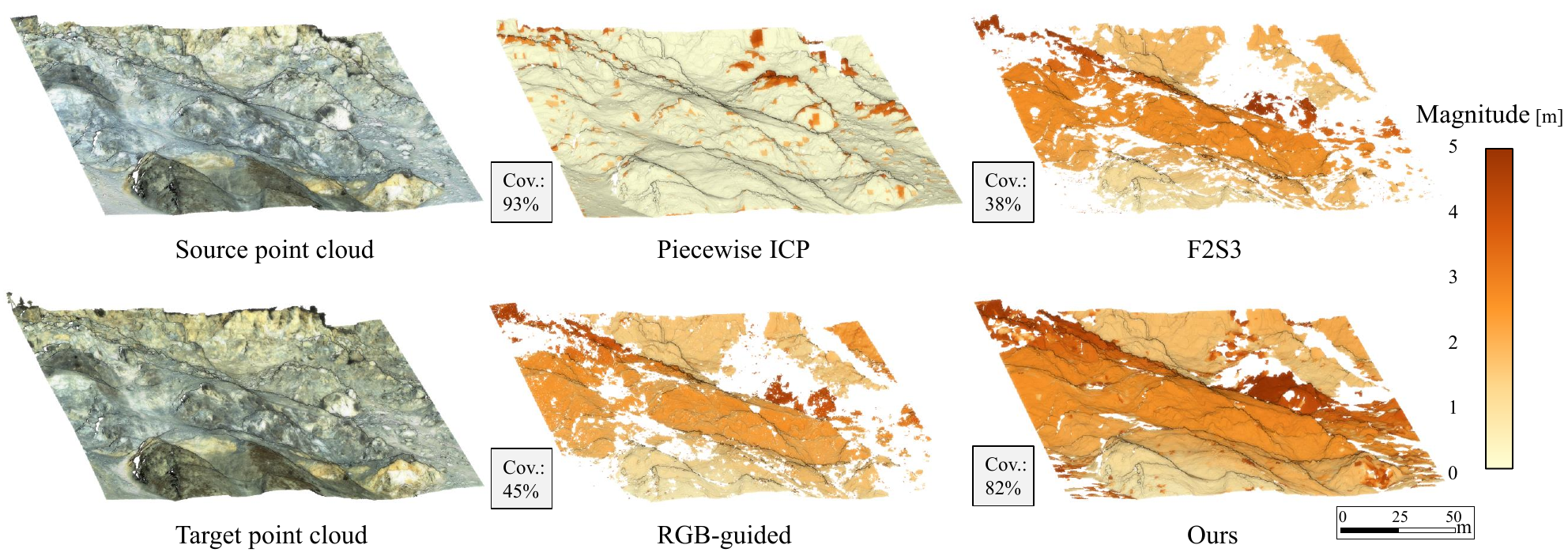}
  \vspace{-1em}
  \caption{Qualitative comparison of displacement magnitudes estimated by different methods on the ROI of the Brienz landslide. Cov. denotes spatial coverage. The results highlight variations among the methods in both spatial coverage and their ability to capture large displacements.}
\label{fig:qualita_diff_methods_brienz}
\end{figure}

\begin{figure}[htb!]
  \centering
  \includegraphics[width=0.9\linewidth]{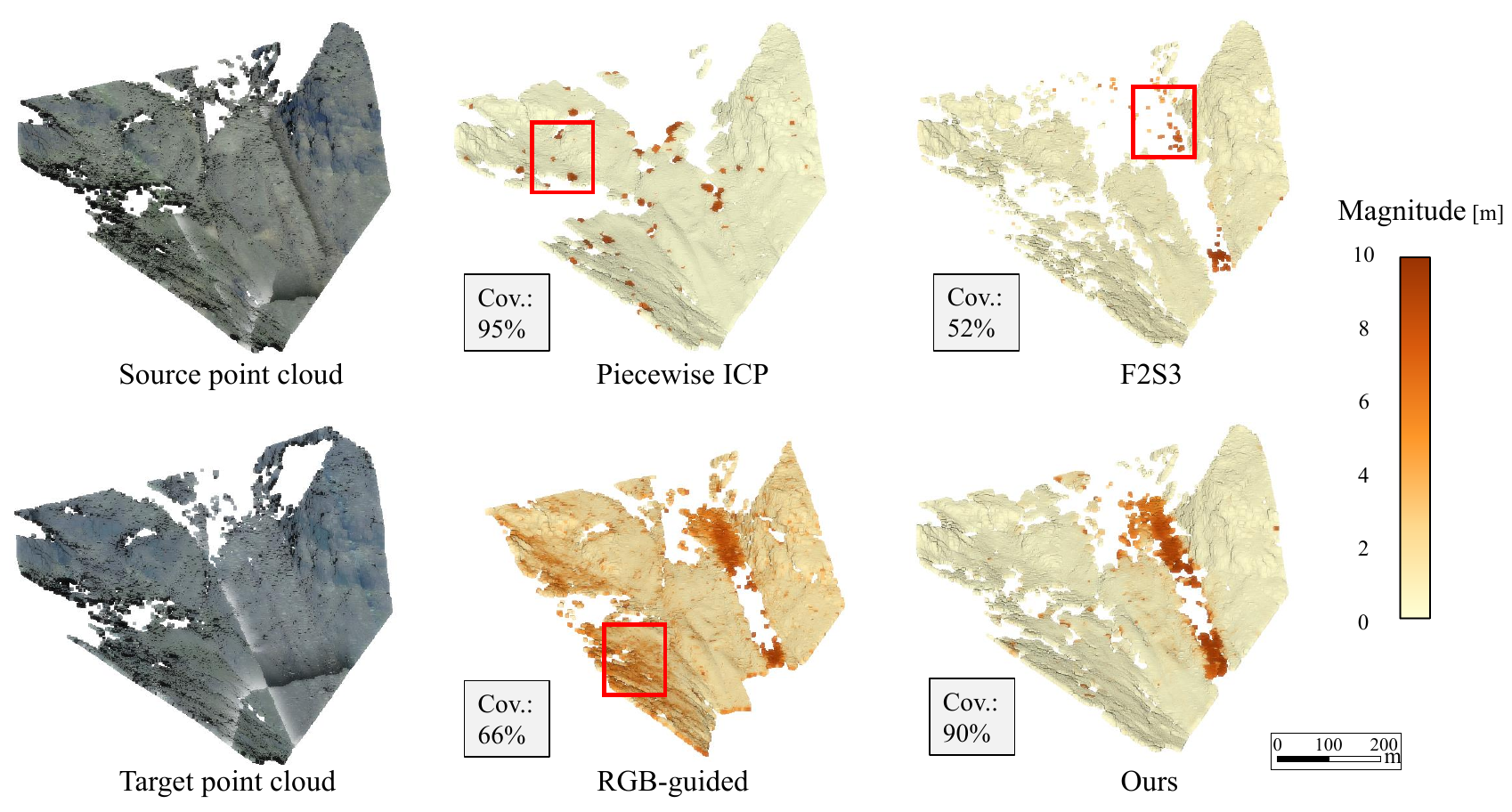}
  \vspace{-1em}
  \caption{Qualitative comparison of displacement magnitudes estimated by different methods on the ROI of the Mattertal landslide, where the results produced by different methods differ in both spatial coverage and local displacement patterns.}
\label{fig:qualita_diff_methods_mattertal}
\end{figure}

\Cref{fig:qualita_diff_methods_brienz} and \Cref{fig:qualita_diff_methods_mattertal} present qualitative comparisons between our method and baseline methods for the ROIs of the Brienz and Mattertal datasets, respectively. 
While Piecewise ICP achieves high spatial coverage (93\% and 95\%) on both ROIs, it underestimates the actual displacements on the Brienz dataset and produces outlier estimates on the Mattertal dataset. In contrast, the other three methods produce results that better reflect the actual displacement trends.
For both ROIs, F2S3 achieves spatial coverage of 38\% and 52\%, and the RGB-guided approach achieves spatial coverage of 45\% and 66\%. These are both much lower than the coverage of 82\% and 90\% achieved by our method. This is attributed to the hierarchical partitioning-based coarse-to-fine estimation and the integrated use of both 3D and 2D information in our method.
Furthermore, the RGB-guided approach produces many noisy estimates on the Mattertal dataset, primarily due to the aforementioned co-registration misalignment errors (see~\Cref{sec:quantitative_results}).
In our method, noisy correspondences from the 2D source are largely suppressed due to the adaptive balance between 3D and 2D information.

\subsection{Effectiveness of the fusion module}\label{sec:discuss_fusion}

\begin{figure}[htb!]
  \centering
  \includegraphics[width=0.6\linewidth]{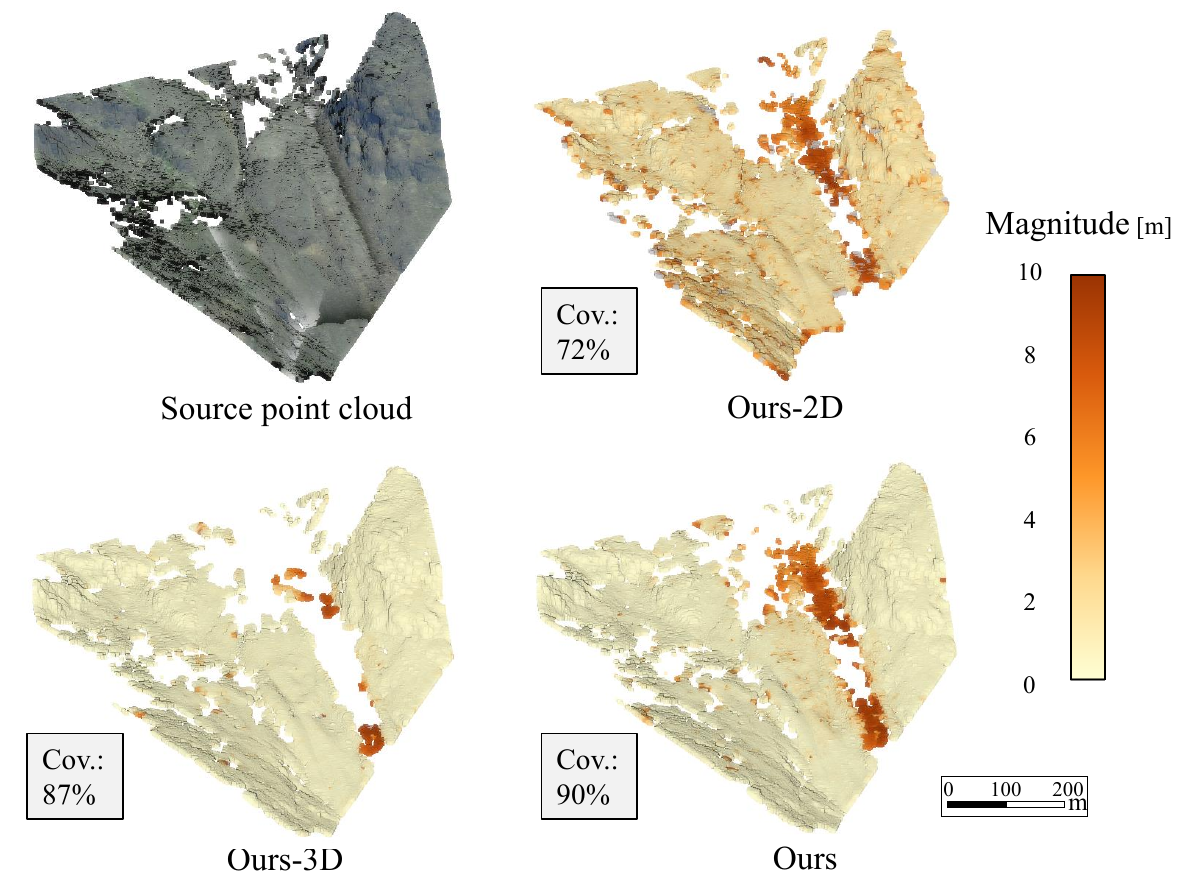}
  \vspace{-1em}
  \caption{Comparison of our method using only 2D information, only 3D information, and both 2D and 3D information on the ROI of the Mattertal landslide, the results of ours achieves the highest spatial coverage while maintaining the high accuracy.}
\label{fig:discuss_fusion_mattertal}
\end{figure}

We evaluate the effectiveness of the fusion module, with qualitative results shown in~\Cref{fig:discuss_fusion_mattertal}. For comparison, we define two ablated variants: Ours-2D, which performs both coarse and fine matching using only RGB image information, and Ours-3D, which relies solely on geometric information from the point cloud. We observe that the RGB-guided method overestimates the displacement magnitudes, consistent with quantitative results in~\Cref{sec:quantitative_results}.
This can be attributed to the co-registration error, which also decreases the quality of 3D displacements estimated by Ours-2D. Conversely, Ours-3D yields more accurate estimates but fails in regions dominated by debris, where geometric features are less discriminative. 
Specifically, at station G1, both Ours (fusion) and Ours-3D have 0.07 m deviations of the estimated displacement magnitudes, while Ours-2D fails to provide a valid estimate. At station G2, Ours-2D exhibits a large deviation of 1.01 m, while Ours-3D fails to provide a valid estimate. Ours has only a 0.28 m deviation. These results highlight the ability of our method to mitigate errors from 2D-only matching while retaining the accuracy of 3D information.
Furthermore, by integrating both 2D RGB and 3D geometry through the fusion module, our method improves spatial coverage to 90\% compared to 87\% of Ours-3D.

\subsection{Impact of multiple levels of partitions}\label{sec:discuss_partition}

\begin{figure}[t!]
  \centering
  \includegraphics[width=0.95\linewidth]{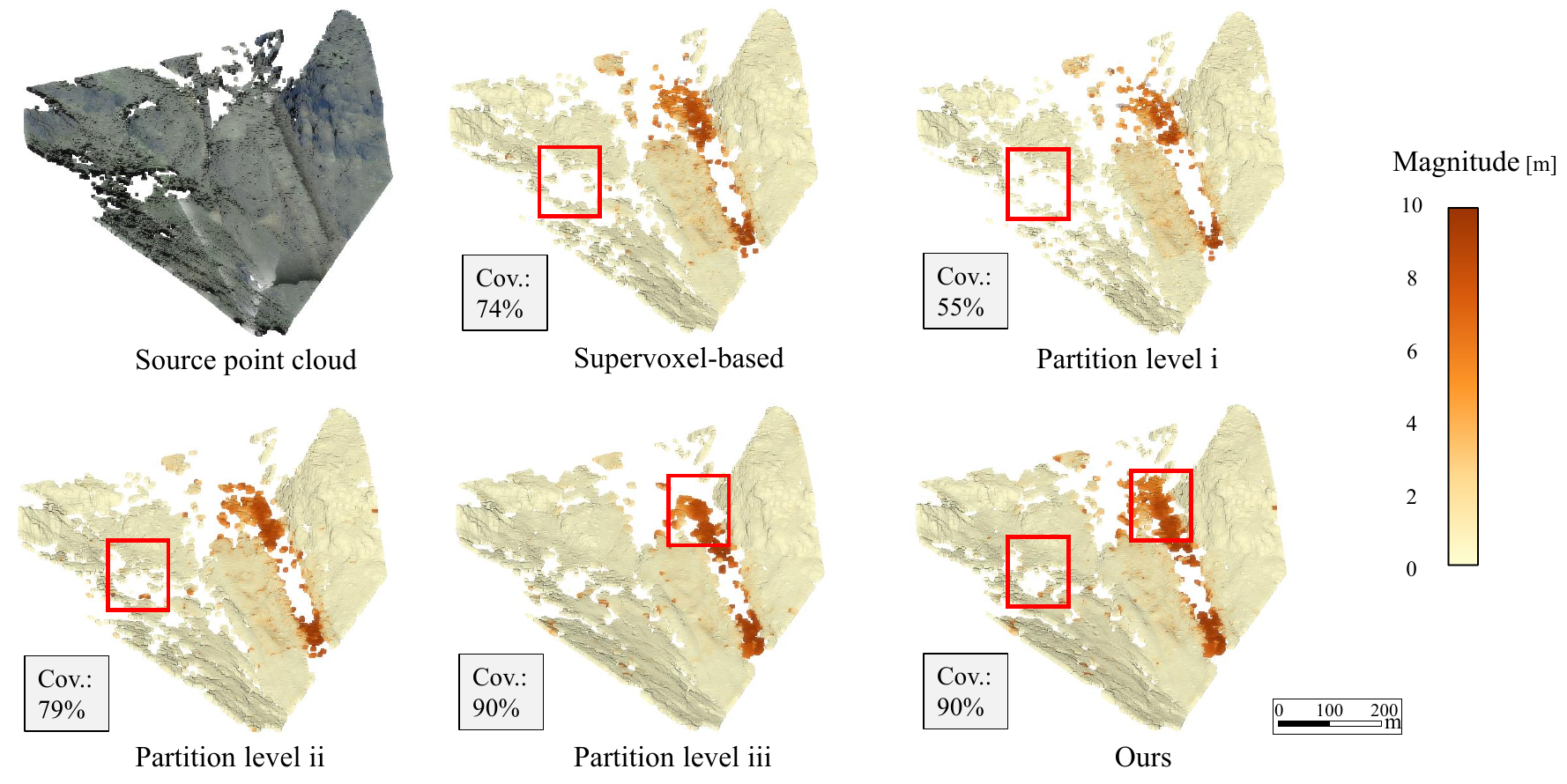}
  \vspace{-1em}
  \caption{Comparison of our method using different partitioning strategies and levels on the ROI of the Mattertal landslide. The optimal configuration (ours) is adopted, as it produces the highest spatial coverage in both quantitative and qualitative evaluations.}
\label{fig:discuss_multi_partition_mattertal}
\end{figure}

We analyze the impact of using different partitioning strategies, with qualitative results shown in \Cref{fig:discuss_multi_partition_mattertal}. 
We apply each of our three hierarchical levels (level i, ii, iii) of partitions independently using the same matching pipeline. 
The supervoxel-based partitioning adopted in F2S3 serves as a baseline and represents a single-level approach. It produces displacements with 74\% spatial coverage, which is higher than our level i (55\%) but lower than level ii (79\%). While level iii offers higher spatial coverage (90\%), it tends to over-smooth displacements by breaking object boundaries. In contrast, our proposed hierarchical method (\cf ours) integrates all three levels incrementally, yielding 90\% spatial coverage while better preserving local structural details.

\subsection{Effectiveness of the refinement module}\label{sec:discuss_refine}

\begin{figure}[htb!]
  \centering
  \includegraphics[width=0.6\linewidth]{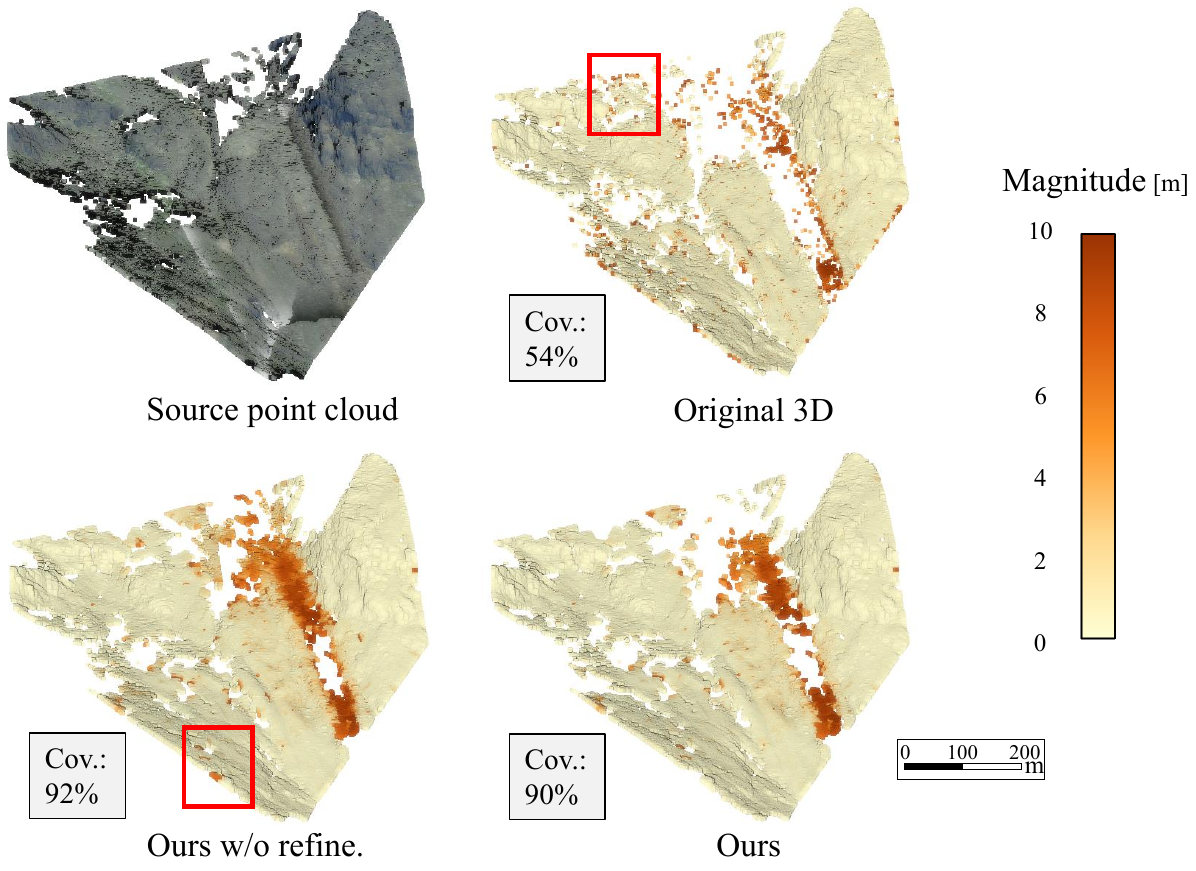}
  \vspace{-1em}
  \caption{Comparison of estimated displacement magnitudes: original 3D output, our method without the refinement module (\cf Ours w/o refine.), and our method with the refinement module (\cf Ours), on the ROI of the Mattertal landslide.}
\label{fig:discuss_refine_module_mattertal}
\end{figure}

We analyze the impact of the refinement module and present the qualitative results in \Cref{fig:discuss_refine_module_mattertal}. We compare three cases: (i) The original 3D result obtained directly from the pretrained deep learning model; (ii) the result obtained using our method without the refinement module; and (iii) the result obtained using our method with the refinement module. The original 3D result has relatively low spatial coverage (54\%) and many noisy estimates, which can be attributed to two factors: 1) The 3D matching fails to produce a sufficient number of matches due to the limitations of geometric features in some regions, where RGB features could contribute, and 2) the absence of the refinement module. When we compare results with and without the refinement module, we find that our method without refinement yields an output with higher spatial coverage (92\%) but more noisy estimates. In contrast, our method with the refinement module effectively eliminates low-quality patch matches, primarily consisting of noisy correspondences, by applying a strict and practical filtering criterion. This improves overall match reliability while maintaining high spatial coverage (90\%) of the output.

\subsection{Runtime analysis}
\label{sec:runtime}

The proposed approach has an overall time complexity of $O(N \log N)$ due to the NN search between 3D points in feature space. We use the GPU implementation of Faiss~\citep{johnson2019billion} for NN search. We further compare the runtime between different approaches, with one image used for both the RGB-guided baseline and our approach. A pair of tiles from the Brienz ROI region (512k source and 571k target points) is used for this analysis. To make a fair comparison, we evaluate all methods on a single GeoForce RTX 3090 Ti with an AMD Ryzen 7 5800x 8-Core Processor. The runtimes for the RGB-guided approach (2D RGB approach), and F2S3 (3D geometry approach) are 81 s and 133 s, respectively. 
Our fusion approach requires 221 s when only partition level~ii is used, which is comparable to the sum of the RGB-guided and F2S3 runtimes. Using all three partition levels increases the runtime to 259 s, indicating that higher spatial coverage can be achieved with only a moderate runtime increase. The runtime of the Piecewise ICP is only 11 s, which demonstrates its high processing efficiency.

\subsection{Limitations and potential improvements}\label{sec:discuss_limitation}

\textbf{Limitations:} Our approach has several limitations. First,
it relies on feature distinctiveness for accurate point-wise or pixel-wise matching. In regions with repetitive geometric structures and weak radiometric textures, \eg, debris-covered surfaces and vegetation regions, or in cases of severe occlusion and topographic disruption caused by large toppling or rotational failures, the method may fail to establish accurate correspondences, resulting in missing displacement estimates.
Second, the current fusion strategy performs coarse matching using 3D point features and 2D pixel features independently, without exploiting deep feature-level fusion. This limitation is partly attributed to the scarcity of annotated, landslide-specific training data.
\textbf{Potential improvements:} 
(i) We adopt the 3D patch-based model from \citep{3d_pretrain21} and the image matching algorithm from \citep{wang2024eloftr}, both of which are commonly used in related applications. While they currently provide strong performance, we believe that future advances in feature learning and image matching may further improve the spatial coverage and accuracy. Fine-tuning the pretrained model with simulated data may be another option.
(ii) Due to the lack of dense ground-truth data for evaluation, one could investigate and evaluate the approach using simulation-based datasets. 
(iii) The method is applied to TLS datasets with images captured by built-in, calibrated cameras. Since these cameras are closely aligned with the scanner in terms of viewpoint, fusion benefits are limited to the feature level. We anticipate that integrating images from different viewpoints, such as stationary cameras~\citep{neyer2016monitoring} or UAV platforms~\citep{uav_platform}, could introduce complementary information at the data level. We leave this exploration for future work.

\section{Conclusion}\label{sec:conclusion}
This paper presents a hierarchical partitioning-based approach for estimating dense 3D DVF in TLS-based landslide monitoring. The method optimizes the matching pipeline in a coarse-to-fine manner using both 3D point clouds and 2D RGB images. By integrating multi-level partition information, we effectively capture structural variations and mitigate the occurrence of noisy estimates while maintaining high spatial coverage. 
Furthermore, rather than requiring annotated landslide-specific training datasets, which are difficult to obtain, the method leverages features extracted from models pretrained on publicly available indoor datasets.
Experiments on two real-world landslide datasets with different displacement patterns validate the effectiveness of the proposed method: it produces displacement estimates with higher spatial coverage than the current state-of-the-art method (F2S3) while maintaining comparable accuracy.
Although this method has been verified for TLS-based landslide monitoring, it can be extended to other types of point clouds and monitoring applications. Future research will focus on exploring the integration of TLS point clouds with RGB images captured from external cameras, such as those mounted on UAVs. This could provide multi-perspective data, enabling landslide monitoring with an enhanced spatial resolution and coverage.

\section*{CRediT authorship contribution statement}

\textbf{Zhaoyi Wang}: Writing -- review \& editing,
Writing -- original draft,
Visualization,
Validation,
Software,
Methodology,
Investigation,
Formal analysis,
Data curation,
Conceptualization.
\textbf{Jemil Avers Butt}: Writing -- review \& editing,
Methodology,
Formal analysis,
Conceptualization.
\textbf{Shengyu Huang}: Methodology,
Conceptualization.
\textbf{Tomislav Medić}: Writing -- review \& editing,
Data curation,
Methodology.
\textbf{Andreas Wieser}: Writing -- review \& editing,
Supervision,
Resources,
Project administration,
Funding acquisition,
Conceptualization.

\section*{Declaration of generative AI and AI-assisted technologies in the writing process}

During the preparation of this work, the authors used ChatGPT in order to assist in refining the readability and language of the manuscript. After using this tool, the authors reviewed and edited the content as needed and take full responsibility for the content of the publication.

\section*{Declaration of competing interest}

The authors declare that they have no known competing financial interests or personal relationships that could have appeared to influence the work reported in this paper.

\section*{Acknowledgments}

{We thank the anonymous reviewers for their constructive comments and suggestions, which helped improve the clarity and quality of this manuscript.
The Brienz dataset was provided by Robert Kenner (Institute for Snow and Avalanche Research), and total station observations were provided by Stefan Schneider (CSD INGENIEURE AG). 
The data collection of the Mattertal dataset was financially supported by the Federal Office for the Environment (FOEN, grant 20.0011.PJ
/6DDBCAD6D) and by the Swiss Geodetic Commission (SGC).
Helena Laasch has shared the code for the Piecewise ICP. The first author was supported by the China Scholarship Council (CSC).

\section*{Data and code availability}

Our example data and source code are publicly available at~\url{https://github.com/gseg-ethz/fusion4landslide}.

\appendix

\section{Deviation patterns along the X- and Z-axis directions}
\label{sec:appendix_xy_deviation}

See~\Cref{fig:quantitative_compare_xz}.

\begin{figure}[H]
  \centering
  \includegraphics[width=0.7\linewidth]{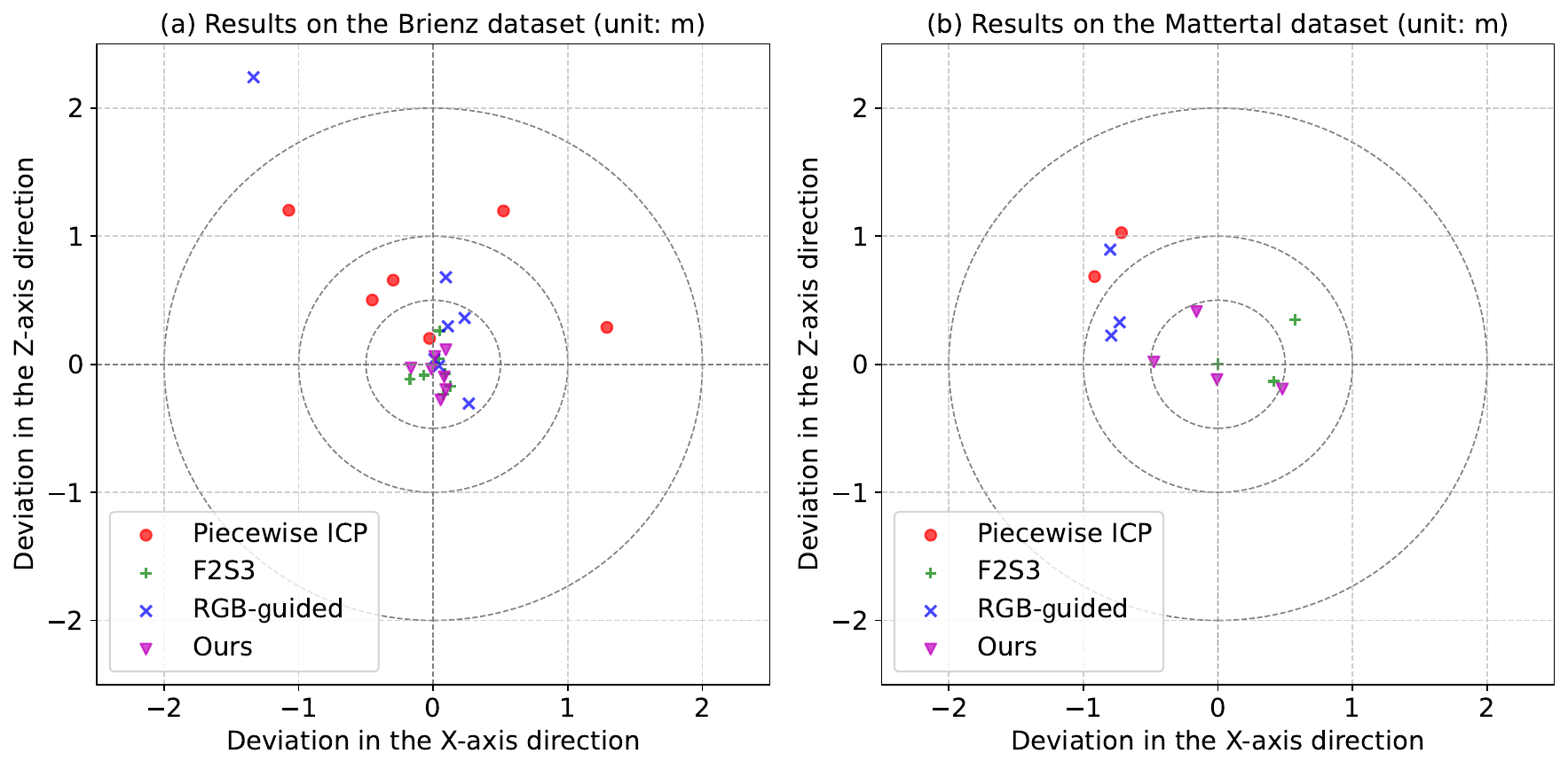}
  \vspace{-1em}
  \caption{Quantitative comparison of estimated displacements from different methods w.r.t. the manual reference, focusing on deviations along the X- and Z-axis directions (\ie, $\Delta D_X$ and $\Delta D_Z$). Dashed circles and grid lines are added to improve readability.}
\label{fig:quantitative_compare_xz}
\end{figure}

\section{Justification of MADD thresholds}\label{sec:madd_sensitive}

For the MADD thresholds (\cf~\Cref{sec:refine}), $\delta_1$ controls the tolerance for geometric deviation within patches, while $\delta_2$ enforces minimum structural consistency between source and target patches. These thresholds were manually tuned according to point cloud resolution and scene scale. Note that $\delta_1$ does not correspond directly to single point spacing, but rather to the mean pairwise distance deviation within a patch. Since patches span structures of several meters, deviations on the order of 1–2 m are acceptable to capture structural consistency.

\begin{table}[h]
  \centering
  \caption{Justification of MADD threshold selection criteria.}
  \label{tab:sensitivity}
  \resizebox{0.8\columnwidth}{!}{
  \begin{tabular}{c|cccc|cccc}
      \toprule
      & \multicolumn{4}{c|}{${|\Delta D_S|}_{\mathrm{Ours - Manual}}$ (m) $\downarrow$} & \multicolumn{4}{c}{Spatial coverage $\uparrow$} \\
      & ${\delta}_2=0.05$ & ${\delta}_2=0.10$ & ${\delta}_2=0.15$ & ${\delta}_2=0.20$
            & ${\delta}_2=0.05$ & ${\delta}_2=0.10$ &${\delta}_2=0.15$ & ${\delta}_2=0.20$ \\
      \midrule
      ${\delta}_1=0.50$ & 0.11 & 0.12 & 0.11 & \textbf{0.09} & 73\% & 74\% & 73\% & 73\% \\
      ${\delta}_1=1.00$ & 0.10 & 0.11 & \textbf{0.09} & 0.10 & 78\% & 78\% & 78\% & 78\% \\
      ${\delta}_1=1.50$ & \textbf{0.09} & \textbf{0.08} & 0.11 & \textbf{0.09} & 81\% & 82\% & 81\% & 81\% \\
      ${\delta}_1=2.00$ & 0.11 & 0.12 & \textbf{0.09} & 0.12 & \textbf{83\%} & \textbf{83\%} & \textbf{83\%} & \textbf{83\%} \\
      \bottomrule
  \end{tabular}
  }
\end{table}

To assess sensitivity, we vary ${\delta}_1 \in [0.50, 1.00, 1.50, 2.00]$ and ${\delta}_2 \in [0.05, 0.10, 0.15, 0.20]$ on the Brienz ROI. For each setting, we report the mean absolute deviation of displacement magnitudes across three available stations (T1, T3, T7), as well as the spatial coverage (see \Cref{tab:sensitivity}). Results show that the mean absolute deviation remains stable across all settings, with fluctuations within 0.04 m. The spatial coverage is largely insensitive to $\delta_2$, while larger $\delta_1$ increases coverage by accepting more matches. We therefore set $\delta_1 = 1.5$ and $\delta_2 = 0.1$ as a balanced configuration in our experiments.

\begin{figure}[htb!]
  \centering
  \includegraphics[width=0.9\linewidth]{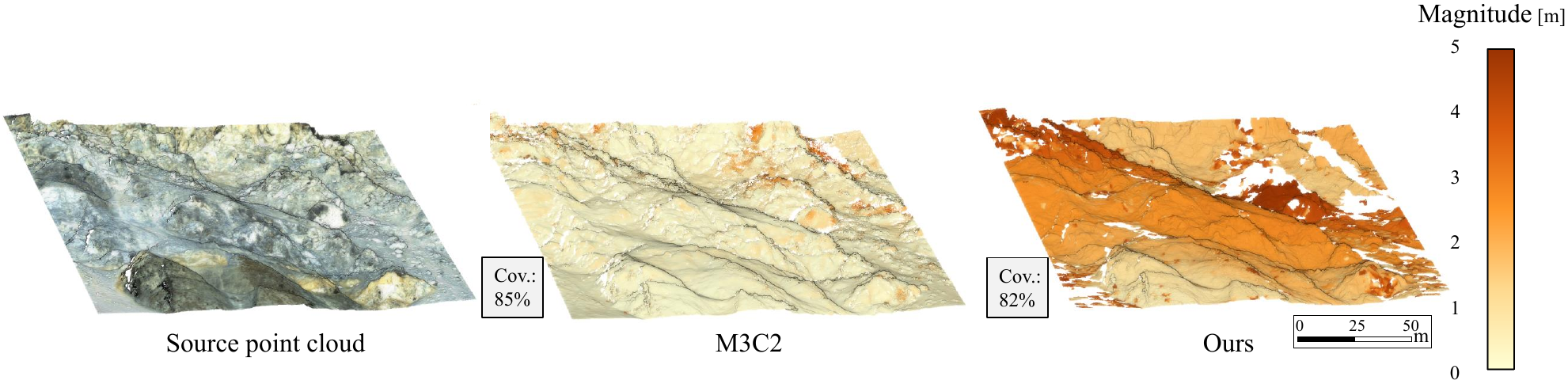}
  \vspace{-1em}
  \caption{Qualitative comparison of displacement magnitudes estimated by M3C2 and our approach on the ROI of the Brienz landslide. In this region, M3C2 yields small displacement magnitudes, whereas our method provides large estimates that have been validated through previous analyses using total station measurements and manually selected reference points (see~\Cref{tab:quantitative_external_compare}), results from TS prisms T1, T3, and T7.}
\label{fig:qualita_m3c2_brienz}
\end{figure}

\begin{figure}[htb!]
  \centering
  \includegraphics[width=0.9\linewidth]{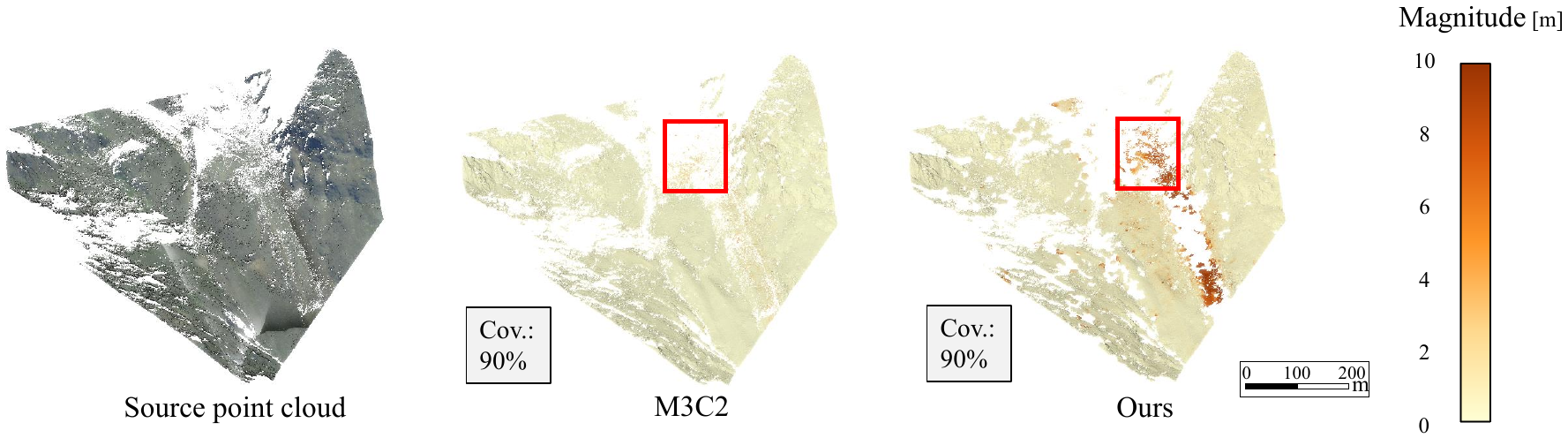}
  \vspace{-1em}
    \caption{Qualitative comparison of displacement magnitudes estimated by M3C2 and our approach on the ROI of the Mattertal landslide. The red rectangle marks an area where M3C2 misses large displacements, while our method provides consistent estimates (see~\Cref{tab:quantitative_external_compare}, results from GNSS stations G1 and G2).}
\label{fig:qualita_m3c2_mattertal}
\end{figure}

\section{Comparison with M3C2 results}

We present comparison results with M3C2 in two ROIs from two landslide in \Cref{fig:qualita_m3c2_brienz} and \Cref{fig:qualita_m3c2_mattertal}, respectively. We implement M3C2 using CloudCompare v2.13.2, and use estimated absolute M3C2 distances for this comparison.
Regarding spatial coverage, M3C2 slightly outperforms our approach on the Brienz ROI by 3\%, while achieving similar spatial coverage on the Mattertal ROI. However, notable deviations are observed in the estimated displacements. This discrepancy arises because M3C2 only estimates distances along a predefined direction, \eg, the LoS direction, rather than full 3D displacements as our method and the selected baselines. 
This observation aligns well with previous work~\citep{f2s3_v2}.
In real-world landslide regions, displacements typically occur along complex surfaces and vary across different parts of the landslide, making it difficult for a single or few directions to represent the overall changes. These results highlight the advantage of feature-based 3D displacement estimation for capturing complex, spatially varying movements in landslide areas.
However, for the same pair of tiles used in~\Cref{sec:runtime}, the computation performed by M3C2 takes only 0.5 s, demonstrating the high efficiency of this CloudCompare plug-in algorithm.

\section{Verification of estimated large displacements}
\label{sec:large_displace}

\begin{figure}[htb!]
  \centering
  \includegraphics[width=0.7\linewidth]{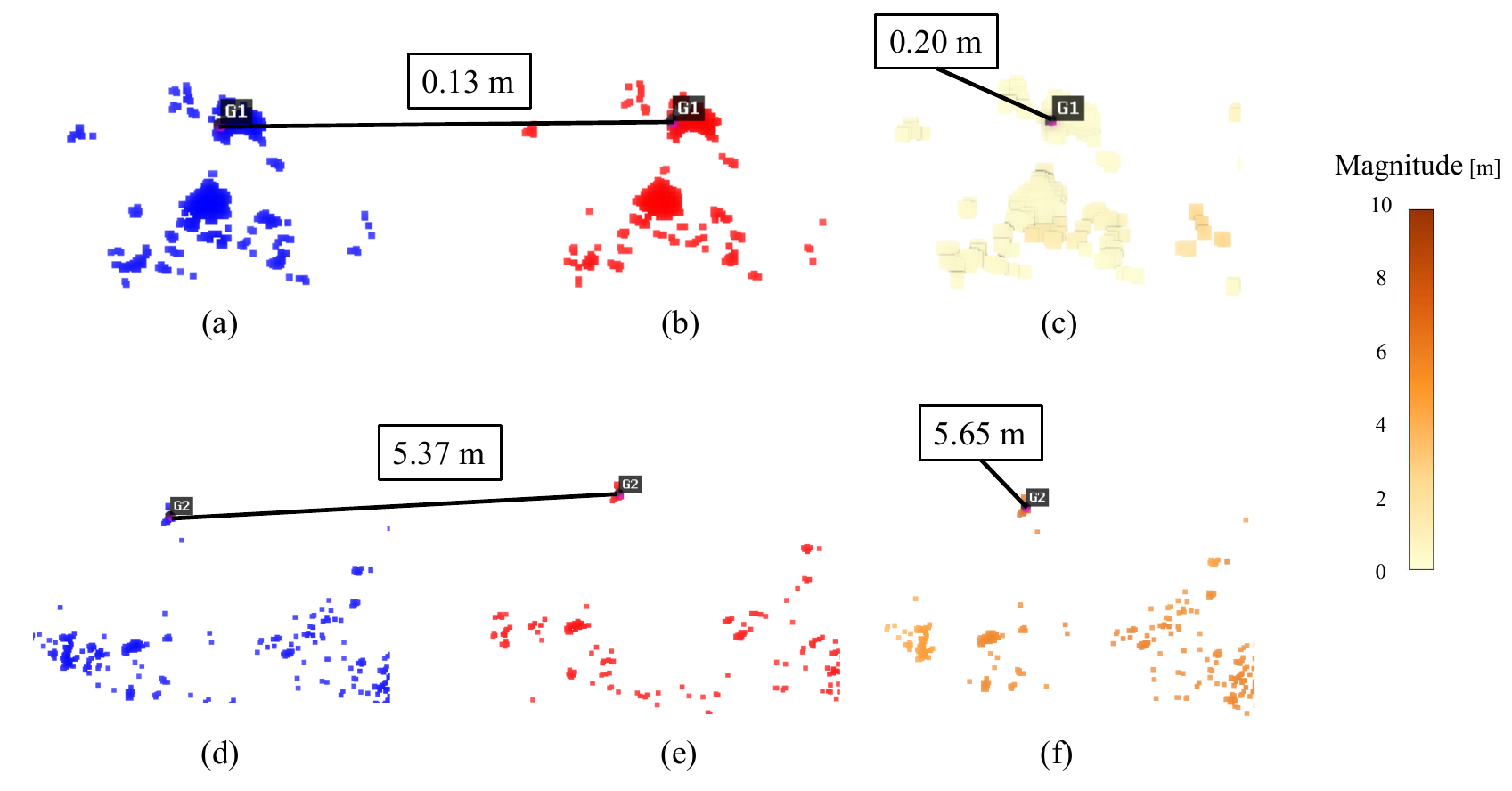}
  \vspace{-1em}
  \caption{Comparison of displacement magnitudes on the Mattertal dataset: (a), (d): first-epoch regions (highlighted in blue); (b), (e): second-epoch regions, and (c), (f): displacements estimated by our method.}
\label{fig:verification}
\end{figure}

In two of the study areas, the detected displacement reaches up to 5.65 m near GNSS station G2. Although such values may appear unusually large, they fall within the range of displacement magnitudes documented for active landslide regions in the Swiss Alps (\eg,~\citep{mattertal,KENNER2025108343,landslide_swiss25}). These studies also discuss the geological and geomechanical context in which such slope movements occur in the same or comparable alpine environments.
It is indeed remarkable that a feature-based approach can capture such large displacements. This indicates that the displacement is mostly a translation and thus the features are preserved. These displacements are captured using a coarse-to-fine feature-based matching strategy that leverages both point cloud geometry and RGB image information (see~\Cref{sec:method}), achieving point-spacing level accuracy. To further illustrate the validity of the large observed displacement,~\Cref{fig:verification} presents a comparison between manually selected point correspondences and the estimates of the proposed approach on the Mattertal dataset.

\bibliography{bib}
\end{document}